\documentclass[10pt,twocolumn,letterpaper]{article}

\usepackage[pagenumbers]{cvpr} 

\usepackage[dvipsnames]{xcolor}

\usepackage[utf8]{inputenc} 
\usepackage[T1]{fontenc}    
\usepackage{amsfonts}       
\usepackage{nicefrac}       
\usepackage{microtype}      

\usepackage{graphicx}
\usepackage{booktabs}

\usepackage{times}
\usepackage{epsfig}
\usepackage{amsmath}
\usepackage{amssymb}
\usepackage{flushend}
\usepackage{tabularx}
\usepackage[fleqn]{nccmath}

\usepackage{pifont}
\usepackage{colortbl}
\usepackage{multirow}
\usepackage{adjustbox}

\usepackage{minitoc}

\newcommand{\smallparagraph}[1]{\medskip\noindent\textbf{#1}~}

\definecolor{MyGreen}{RGB}{0, 180, 0}
\definecolor{MyRed}{RGB}{180, 0, 0}
\definecolor{MyBlue}{RGB}{30, 0, 180}

\newcommand{\cmark}{{\textcolor{MyGreen}{\ding{51}}}}%
\newcommand{\xmark}{{\textcolor{MyRed}{\ding{55}}}}%

\newcommand{\ours}{\text{UNIT}\xspace}

\newcolumntype{R}[2]{%
    >{\adjustbox{angle=#1,lap=1.3\width-(#2)}\bgroup}%
    l%
    <{\egroup}%
}

\usepackage{dsfont}
\usepackage{etoolbox}

\newif\ifshowedits

\newcommand{\addeditor}[3]{%
  \definecolor{#1color}{rgb}{#3}
  \expandafter\newcommand\csname #1\endcsname[1]{%
  \ifshowedits
    {\color{#1color} ##1}%
  \else
    {##1}%
  \fi
  }%
  \expandafter\newcommand\csname #1rmk\endcsname[1]{%
  \ifshowedits
    {\color{#1color} {\bf [#2: ##1]}}
  \fi
  }%
  \expandafter\newcommand\csname #1rpl\endcsname[2]{%
  \ifshowedits
    {\color{#1color} ##1 \sout{##2}}
  \else
    {##1}
  \fi
  }%
}

\newcommand{\createtextvar}[1]{
  \expandafter\newcommand\csname #1\endcsname{%
  {\text{#1}}
}%
}
\newcommand{\mycomment}[1]{}

\DeclareMathOperator*{\argmax}{arg\,max}

\graphicspath{ {./figures/} } 

\definecolor{cvprblue}{rgb}{0.21,0.49,0.74}
\usepackage[pagebackref,breaklinks,colorlinks,citecolor=cvprblue]{hyperref}

\addeditor{vincent}{VL}{0.0, 0.5, 0.0}
\addeditor{gilles}{GP}{0.0, 0.5, 0.9}
\addeditor{renaud}{RM}{0.95, 0.55, 0.15}
\addeditor{alex}{AB}{0.6, 0.4, 1.0}
\addeditor{corentin}{CS}{1.0, 0.0, 1.0}
\showeditstrue
\showeditsfalse

\def\paperID{246} 
\def\confName{3DV\xspace}
\def\confYear{2025\xspace}

\title{UNIT: Unsupervised Online Instance Segmentation through Time}

\author{
Corentin Sautier$^{1,2}$ \quad
Gilles Puy$^{2}$ \quad
Alexandre Boulch$^{2}$ \quad
Renaud Marlet$^{1,2}$ \quad
Vincent Lepetit$^{1}$ \quad
\and
\and
\large
\hspace{-3mm} \textsuperscript{1}LIGM, Ecole des Ponts, Univ Gustave Eiffel, CNRS,  France
\hspace{1mm} \textsuperscript{2}Valeo.ai, France 
}

\begin{document}
\maketitle

\begin{abstract}
Online object segmentation and tracking in Lidar point clouds enables autonomous agents to understand their surroundings and make safe decisions. Unfortunately, manual annotations for these tasks are prohibitively costly. We tackle this problem with the task of class-agnostic unsupervised online instance segmentation and tracking.
To that end, we leverage an instance segmentation backbone and propose a new training recipe that enables the online tracking of objects. Our network is trained on pseudo-labels, eliminating the need for manual annotations. We conduct an evaluation using metrics adapted for temporal instance segmentation. Computing these metrics requires temporally-consistent instance labels. When unavailable, we construct these labels using the available 3D bounding boxes and semantic labels in the dataset. We compare our method against strong baselines and demonstrate its superiority across two different outdoor Lidar datasets.
Project page: \href{https://csautier.github.io/unit}{csautier.github.io/unit}
\end{abstract}


\section{Introduction}

\gilles{Being able to segment, classify, and track all object instances in sequences of point clouds acquired by Lidars is critical for many applications, including autonomous driving. This task,}
called 
(Multi-Object) Panoptic Tracking~\cite{pt,mopt}, can be seen as a temporal version of panoptic segmentation. 

Unfortunately, the cost of annotating point cloud data for this task can be enormous~\cite{semantickitti}. Moreover, to ensure high performance \renaud{and robustness}, a dataset should contain enough examples of rare objects \renaud{(e.g., one-wheelers, electric scooters)} or \renaud{hard settings} 
\renaud{(e.g., adverserial weather)}, increasing the cost of annotation further. 
\renaud{For systems with an international scope, such as autonomous driving, there is also a need to address domain shifts and cover different object distributions (e.g., shape variations \cite{wang2020train}) and local specificities (e.g., rickshaws).}
\gilles{One way to reduce the need for annotations is to rely on self-supervised~\cite{slidr,also,bevcontrast,tarl,segcontrast,scalr} or unsupervised learning techniques~\cite{oyster,3duis,semoli,liso}. }

\begin{figure*}[t]
    \centering 
    \vspace*{-2mm}
    \newcolumntype{a}{>{\columncolor{blue!10}}c}
    \setlength{\tabcolsep}{2pt}
    \begin{tabular}{@{}cccac@{}}
        \rotatebox{90}{SemanticKITTI} &
        \includegraphics[width=0.238\linewidth, trim={0 2.4cm 0 2cm},clip]{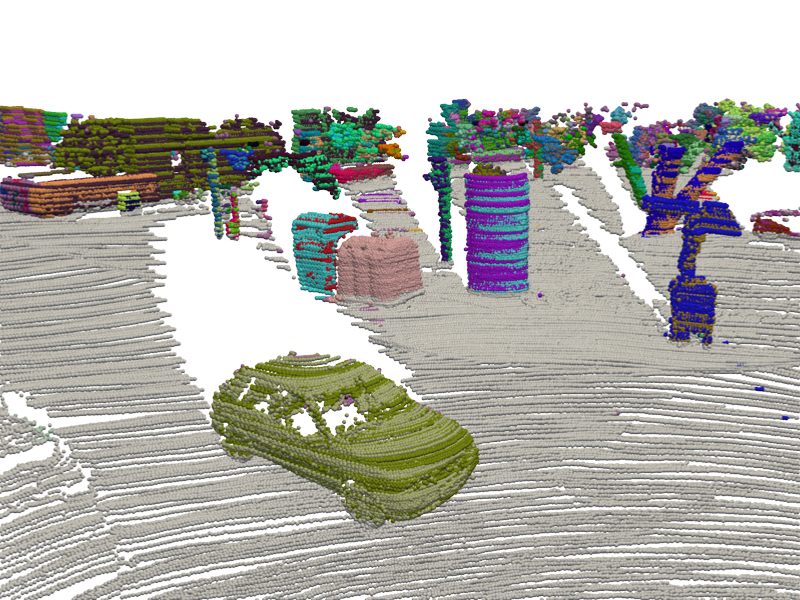}&
        \includegraphics[width=0.238\linewidth, trim={0 2.4cm 0 2cm},clip]{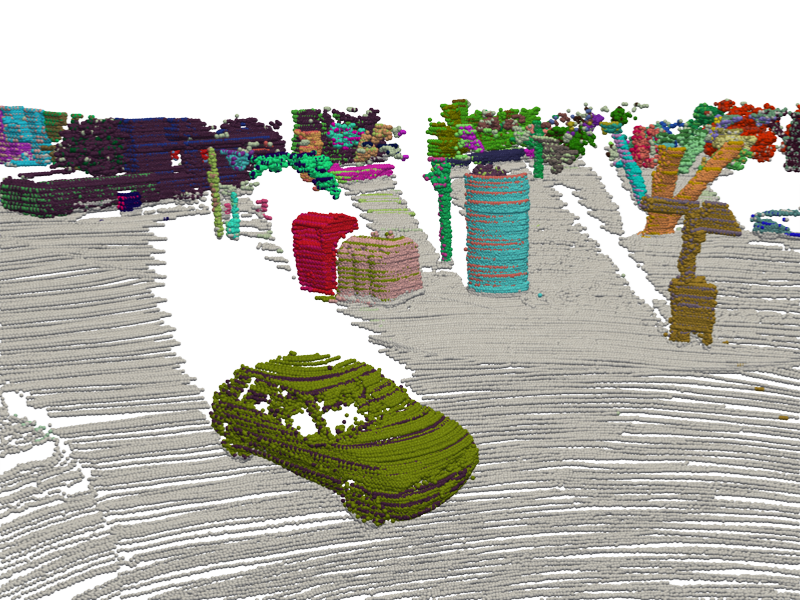}&
        \includegraphics[width=0.238\linewidth, trim={0 2.4cm 0 2cm},clip]{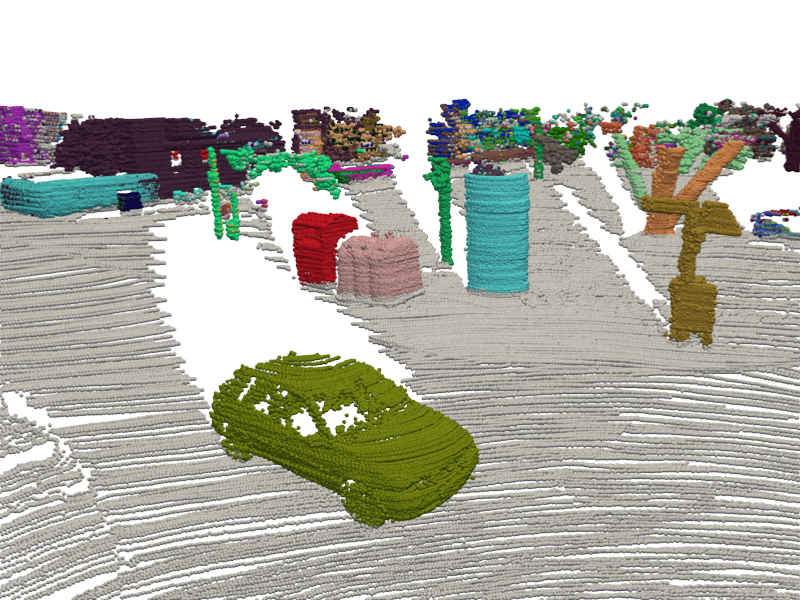}&
        \includegraphics[width=0.238\linewidth, trim={0 2.4cm 0 2cm},clip]{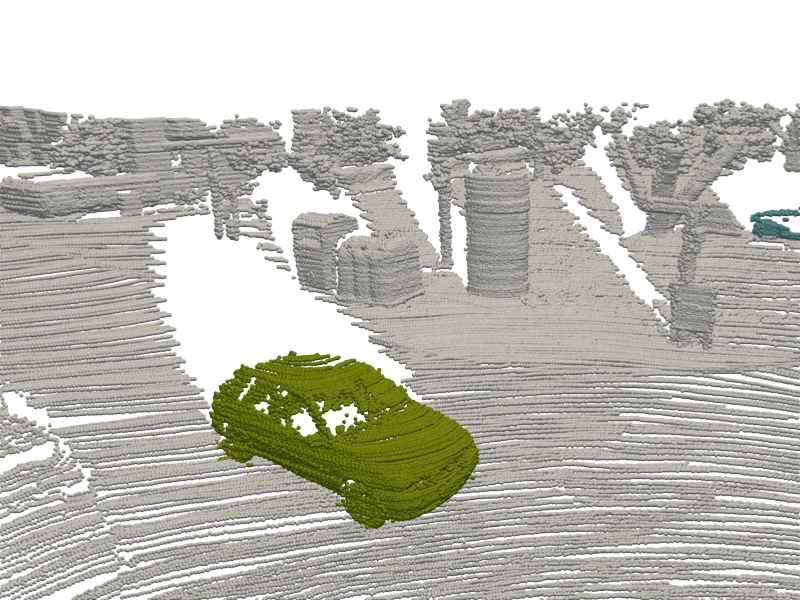}
        \\
        \rotatebox{90}{\hspace{3mm}PandaSet-GT} &
        \includegraphics[width=0.238\linewidth, trim={0 4cm 0 0},clip]{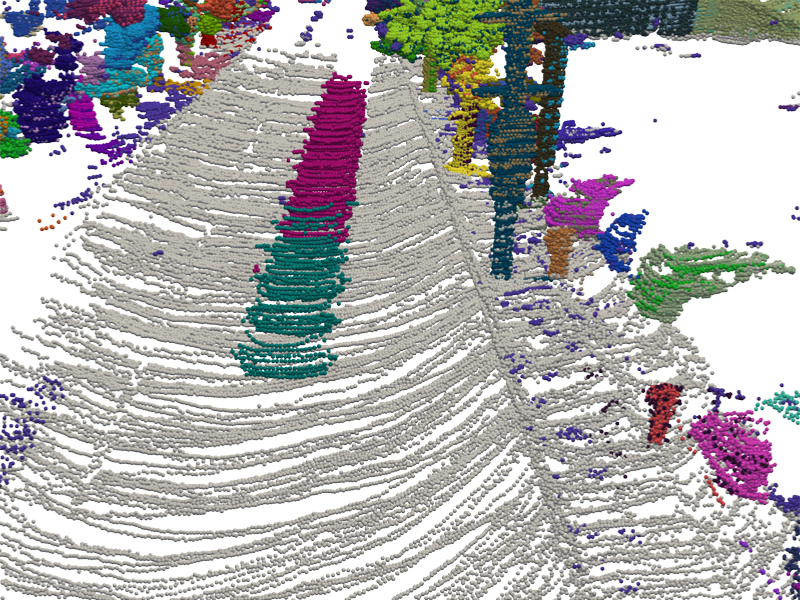}&
        \includegraphics[width=0.238\linewidth, trim={0 4cm 0 0},clip]{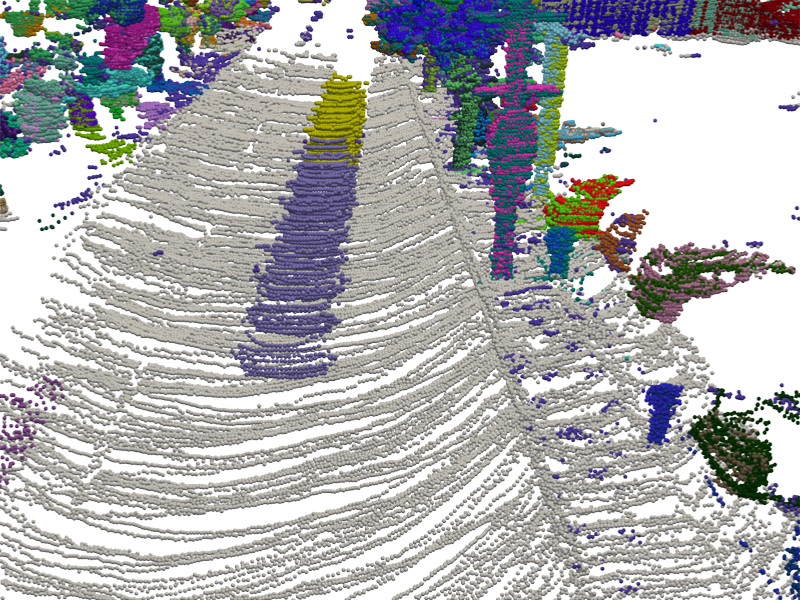}&
        \includegraphics[width=0.238\linewidth, trim={0 4cm 0 0},clip]{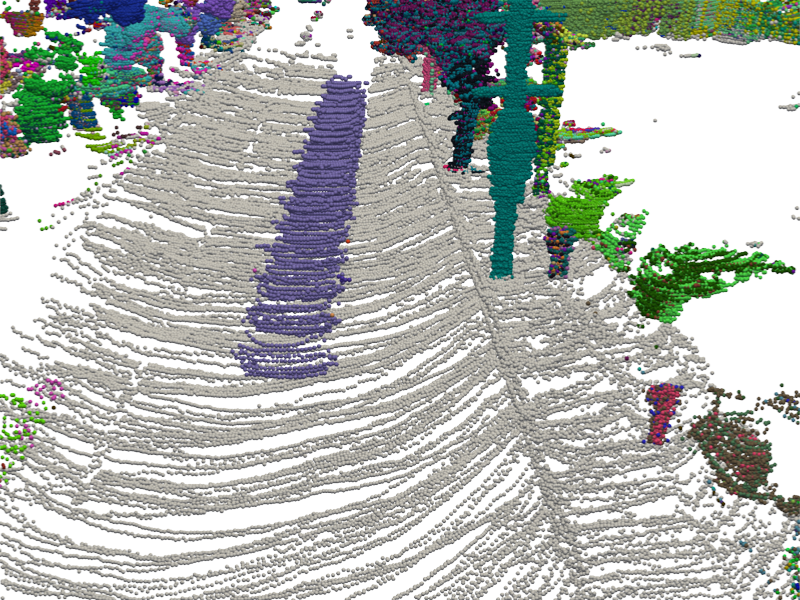}&
        \includegraphics[width=0.238\linewidth, trim={0 4cm 0 0},clip]{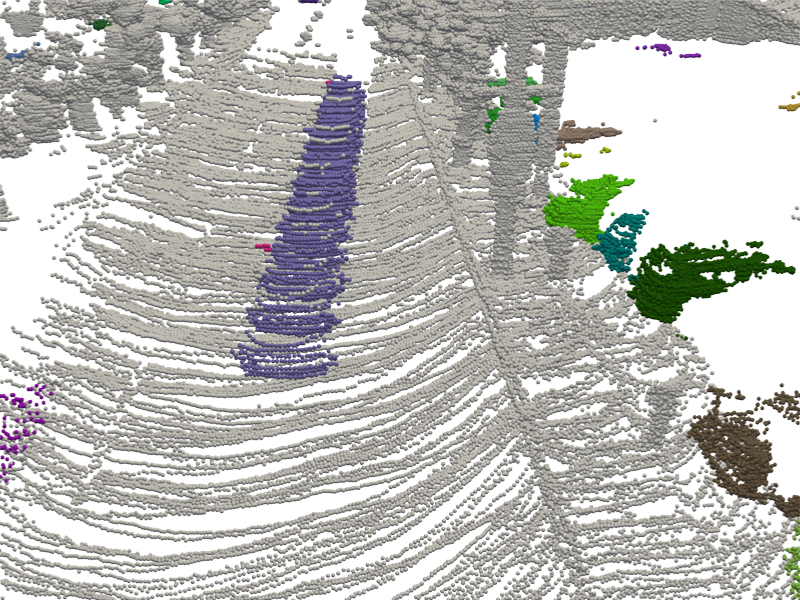}
        \\
        &
        \small TARL \renaud{offline} segments & 
        \small \renaud{Our} \renaud{offline} 4D segments & 
        \small \renaud{Overlay of our online segments} & 
        \small Ground truth
        \\
        &
        \small \renaud{(TARL-Seg)} & 
        \small \renaud{(4D-Seg)} & 
        \small (\ours) & 
        \small 
    \end{tabular}
    \vspace*{-3mm}
    \caption{We generate pseudo-labels of instances across time in Lidar scans \renaud{(4D-Seg)}, which we use to train an online segmenter \renaud{(\ours)} that assigns to each 3D point an instance ID consistent over time.
    \renaud{We show aggregations of scans over time. The sample scene from Semantic- KITTI, with the car in the foreground, is mostly static, while the scene from PandaSet-GT is dynamic, with the trace of a moving vehicle. \newline\mbox{}\hspace*{12pt} Column~1 shows TARL \cite{tarl} spatio-temporal segments. Column~2 shows our improved 4D segments. Column~3 shows an aggregation of our online labeling of successive scans (our input is \emph{not} an aggregated scan). Column~4 shows the ground truth. Segment colors are random. \newline\mbox{}\hspace*{12pt}}
    Note that we obtain more labels than in the ground truth \renaud{as our class-agnostic segmentation  include all objects and stuff, such as trees or buildings}. Also, while our architecture performs online, 
    it outperforms the offline spatio-temporal clustering that we \renaud{use for training.}} 
	\label{fig:renderings}
\end{figure*}


In this work, we introduce a new \renaud{fully unsupervised} method, which we call UNIT, to segment class-agnostic object instances and track them over time, 
regardless of their nature or semantics. 
\vincent{More exactly, our goal is to densely segment an input point cloud so that each input point is assigned to one \renaud{and only one} instance ID.}

As illustrated in \cref{fig:renderings}, given unlabeled Lidar sequences, we \renaud{first} obtain 
pseudo 4D segments of objects by using a spatio-temporal clustering. Our clustering is inspired by SegContrast~\cite{segcontrast} and TARL~\cite{tarl}, but we improve these methods to obtain clusters over a longer time window.
This gives us 
4D segments of objects present in the Lidar sequences. We then use these segments as pseudo-labels to train a novel auto-regressive architecture, which, at inference, identifies and track objects in new Lidar sequences by being applied to consecutive \renaud{scans.} 


To evaluate our method, we test it
\alex{on two \renaud{main} datasets, SemanticKITTI \cite{semantickitti} and PandaSet\renaud{-GT} \cite{pandaset},} 
which exhibit different \renaud{point} 
densities, \renaud{scan} 
patterns and 
environments. We use several metrics to analyze both the segmentation and temporal tracking quality of our method. \corentin{We compare our results against strong baselines as well as temporally-extended versions of 3D instance segmentation techniques.} 
\gilles{Finally, we also show the potential of \ours as a self-supervised pretraining technique for semantic segmentation.}

\section{Related work}
\label{sec:related}


\gilles{\paragraph{Object detection/segmentation with transformers.} Our network architecture relies on a transformer decoder similar to the one used in DETR \cite{detr} for object detection, or in \cite{maskformer, mask3d, maskpls} for object segmentation. A set of learnable object queries are provided at the input of a transformer decoder, which is in charge of detecting/segmenting objects in the image or point cloud of interest thanks to a series of cross-attention layers between the object queries and pixel or point features. The transformer is trained so that each query attends to at most one object instance and each object instance is matched to a single query. By default, these methods do not allow objects to be tracked over time.}


\paragraph{Object detection and tracking with transformers.}
\gilles{To detect and track objects in videos, \cite{motr,trackformer,transtrack} extend DETR~\cite{detr} using two sets of object queries: a first set of queries is in charge of tracking discovered objects in past scans, while a second one is in charge of discovering new objects. In contrast, we propose an architecture \renaud{that detects and tracks objects}
with a single set of queries.}


\paragraph{Unsupervised instance segmentation.}
UnScene3D \cite{unscene3d} 
\gilles{performs} 
unsupervised instance segmentation \gilles{in dense point clouds acquired in static indoor scenes. A Mask3D model \cite{mask3d} is trained using pseudo-masks obtained by leveraging self-supervised features and a graph-cut algorithm. Because of the \renaud{static} setting considered, UnScene3D does not require any temporal consistency in predictions.}

SegContrast~\cite{segcontrast} and TARL~\cite{tarl} are self-supervised methods whose pretext tasks rely on pseudo-masks of objects. SegContrast uses a RANSAC~\cite{ransac} algorithm to remove the road plan of Lidar scans, and then a DBSCAN~clustering \cite{dbscan} to obtain pseudo-segments of instances. TARL extends SegContrast by \gilles{working on accumulated scans}, replacing RANSAC with Patchwork~\cite{patchwork}, a better-performing ground segmentation method, and using HDBSCAN instead of DBSCAN~\cite{hdbscan}, which gives better segments especially in low-density regions of the point cloud. 
We compute our pseudo-segments by building upon TARL (see \cref{sec:clustering}). 

3DUIS~\cite{3duis} uses a self-supervised model trained with SegContrast to define affinity in a graph between points, and 
a graph-cut algorithm to get object instances. 
\cite{3duis}~also propose the $\mathrm{S_{assoc}}$ metric for evaluating unsupervised segments. 

\gilles{AutoInst~\cite{autoinst}, like 3DUIS, obtains object segments using a graph-cut algorithm. This graph is built by exploiting TARL \cite{tarl} point features (possibly complemented with DINOv2 \cite{dinov2} features) instead of SegContrast \cite{segcontrast} features. A notable difference between AutoInst and 3DUIS is that AutoInst operates on accumulated scans instead of single scans. These segments are then used as pseudo-labels to train a MaskPLS \cite{maskpls} model, which improves the quality of \renaud{predicted} 
segments. This MaskPLS model, like the graph-cut algorithm, operates on accumulated scans. Unlike AutoInst, \ours operates on single scans instead of accumulated scans, in a online fashion. At inference time, contrary to AutoInst, \ours does not need odometry information for accumulation. \ours also does not need to leverage pretrained models to obtain pseudo-labeled objects segments. 
\renaud{Besides, AutoInst, only evaluates on SemanticKITTI where sequences 01 and 04 are removed from the training set because they predominantly consist of dynamic instances. In contrast, with \ours, there are no requirements limiting the number of dynamic instances for training. Last, the code of AutoInst is not publicly available, which prevents full-fledged comparisons.}
}

OYSTER~\cite{oyster}, SeMoLi~\cite{semoli} and LISO~\cite{liso} use motion cues to find moving objects and regress boxes around them, \renaud{which are used} to train an object detector unsupervisedly. While some aspects are similar to our work, they focus on moving objects and object detection, while we target all objects and instance segmentation. Besides, they do not use a temporal architecture.

\begin{figure*}[t]
    \centering
    \def\svgwidth{\linewidth}
    \input{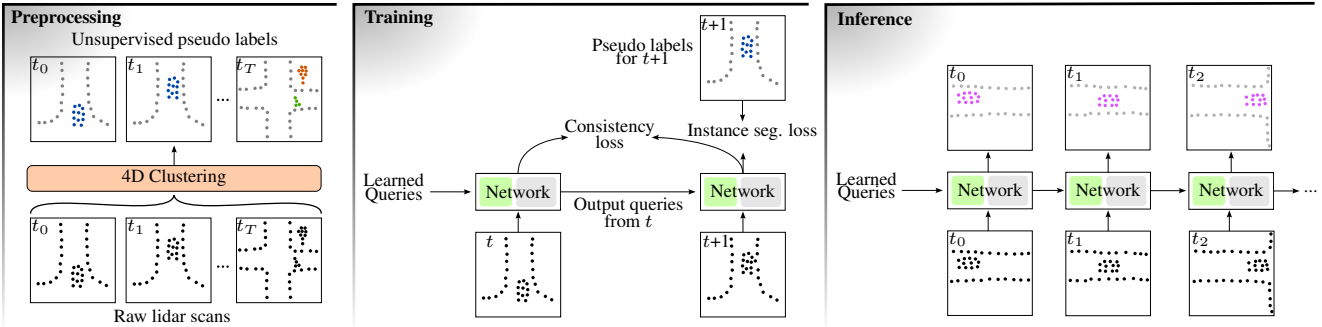}
    \caption{\textbf{Method overview.} 
    Given unlabeled Lidar scans, we create offline pseudo-labels by spatio-temporal clustering \renaud{(preprocessing)}. We then use these pseudo-labels to train an auto-regressive network. 
    At inference, we apply this network 
    to \renaud{successive} scans as they come.
    } 
    \label{fig:method_overview}
\end{figure*}


\section{Method}
\label{sec:method}



\cref{fig:method_overview} provides an overview of our approach: given a sequence of Lidar scans, we train an online instance segmenter that labels each 3D point with an ID for the object to which it belongs, while maintaining consistent IDs over time: two points from different scans sharing the same ID belongs to the same instance. The online, temporal processing is performed by a novel auto-regressive architecture that successively takes 
\renaud{individual} Lidar scans as input and labels points as scans come in \renaud{(without knowing the future)}. 

We train this network without any human supervision. To that end, we first create pseudo-labels on 
\renaud{unannotated} Lidar sequences using a spatio-temporal clustering.
As it is unsupervised and incorporates very few priors beyond 4D consistency, our method makes no distinction between ``thing'' objects (e.g., cars, pedestrians) and ``stuff'' objects (e.g., vegetation, buildings), whose separation may anyway vary from one dataset to another (e.g., for traffic signs). 

We detail each step of our approach below.

\subsection{Pseudo-labeling via spatio-temporal  clustering}
\label{sec:clustering}

\smallparagraph{Road segmentation.} As experimentally verified in prior work~\cite{segcontrast,tarl,oyster}, a good strategy for segmenting objects in automotive Lidar scans is to first remove the ground, as it makes the objects stand out as isolated clusters that are more easily separable.
As the ground can often be assimilated to a large, mostly horizontal plane, RANSAC~\cite{ransac} can be used to separate this plane from the rest of the scan. However, as they take into account the specific patterns of Lidar point clouds, recent alternative methods~\cite{patchwork,patchworkpp} obtain better performance on ground segmentation, while remaining unsupervised.
    
We rely on these methods to segment the ground.
To find the ground points, we used Patchwork~\cite{patchwork} on SemanticKITTI (as in TARL~\cite{tarl}), and Patchwork++~\cite{patchworkpp} 
\alex{on PandaSet-GT.}
Both methods need to be applied scan-wise on each Lidar scan of the training set.

\smallparagraph{Spatio-temporal segmentation.} 
The authors of TARL~\cite{tarl} demonstrate that applying HDBSCAN~\cite{hdbscan} on temporally accumulated scans enables the extraction of spatially and temporally coherent object instances, even for dynamic objects. Yet, the complexity of HDBSCAN restricts the application of this clustering to a short time window. In practice, TARL \cite{tarl} aggregates 12 successive scans of SemanticKITTI, and thus provides pseudo-labels over a temporal window of 1.2\,s as the Lidar operates at 10\,Hz.
To increase the temporal context window, we apply a voxel grid sampling to heavily reduce the number of points to cluster after temporal aggregation.
More details are given in the supplementary material. With this process, we are able to obtain time-coherent clusters over 40 aggregated scans. We will see in \cref{sec:experiments} (\cref{table:results_sk}) that our clusters have the same quality as the clusters obtained in~\cite{tarl} when evaluated on a per-scan basis, but that their temporal consistency is of a much better quality.

\subsection{Online instance segmentation network}
\label{sec:network}

Our \renaud{segmenter} architecture is based on Mask3D~\cite{mask3d}, which \renaud{itself} is an adaptation of MaskFormer~\cite{maskformer} for 3D point clouds. We recall below how these methods work (using object-based queries) and explain what modifications we bring to make them work in a class-agnostic and \renaud{temporal} (online) setting. \renaud{More details 
are provided in appendix.}

\smallparagraph{Class-agnostic object segmentation.} Mask3D uses a transformer decoder to segment objects. \renaud{This} 
decoder attends to point features and produces one embedding for each input query, with a fixed number of queries $|Q|$. 
Each query $q {\,\in\,} Q$ 
is in charge of segmenting one object via a mask module: the scalar product between each query embedding and all point features provide an instance heatmap, which can be transformed into a object-wise binary mask, e.g., via thresholding. 
The modifications we bring to Mask3D are threefold.

First, we have to take into account that, in our unsupervised setting, we have no access to the object classes. We therefore do not use any semantic classification head on top of the query embeddings.

Second, to densely segment the whole input point cloud, each input point must be assigned to a mask ID, and thus to a query. 
To do so, at inference, after computing all scalar products $\langle \text{feat}(p), q\rangle$ between the point features and the query embeddings, 
each point \renaud{$p$ is assigned} the ID of the query that responds the most to the point features:
\begin{align}
    \text{ID}(p) &= \text{ID}(\text{query}(p)) \label{eq:pointID}
    \\
    \text{query}(p) &= \argmax_{q \in Q} \langle \text{feat}(p), q\rangle \> .
    \label{eq:pointquery}
\end{align}
\renaud{(There is no thresholding here as it could leave points without any assigned ID.)} \gilles{Queries to which at least one point is assigned are called active queries.} During training, this hard assignment is not applied and the loss is directly computed on the results of the scalar products (cf.\ \cref{sec:matching}).

Finally, we wish to adapt the queries to follow moving objects. To that end, we consider an auto-regressive architecture where the queries at network output are used again as input to the next network inference (see below).

\begin{figure}
    \centering
    \def\svgwidth{0.75\linewidth}    
    \input{figures/network}
    \caption{Mask3D \cite{mask3d} network architecture. The 
    queries are trained to \renaud{favor} attend\renaud{ing} to at most one object in the Lidar point cloud (\cref{sec:matching}).
    Output point instances are defined from the most responding queries (\cref{eq:pointID,eq:pointquery}).
    \newline\mbox{}\hspace*{12pt}
    We leverage and adapt this architecture for online segmentation and tracking of objects in Lidar sequences \renaud{(cf.\ \cref{sec:network}). To that end, the output queries are input to another instance of the network, in an auto-regressive manner (see \cref{fig:method_overview}).}}
    \label{fig:mask3d}
\end{figure}

\smallparagraph{Auto-regressive extension.}
Mask3D's transformer decoder refines the object queries progressively by attending at point features output at different resolutions (cf.\ \cref{fig:mask3d}). After each refinement \renaud{step}, each query increasingly specializes to a specific object.

To allow the tracking of each instance through time, we inject the query embedding at time $t$ (output of the transformer decoder) as input to the transformer decoder at time $t$+1. As the encoder and decoder share the same weights at each time step, each query at time $t$+1 should therefore continue to attend to the same object that it was already attending to at time~$t$. At least in principle, this architecture enables the tracking of objects over an infinite time window. \renaud{This recursion is illustrated in \cref{fig:method_overview} (training, inference).}

\renaud{As input to the first network application of this auto-regressive scheme \renaud{(see input queries at bottom of \cref{fig:mask3d})}, we use fully-learnable initial queries. This flexibility, which is in line with \cite{detr, maskformer, mask2former} but unlike the non-parametric queries of \cite{3dter,mask3d}, prepare for the recursive use of the queries.}


\subsection{Training loss}
\label{sec:matching}

Our training loss is made of two main terms. The first term is a scan-wise training loss, similar to the one used in Mask3D~\cite{mask3d}. We detail its construction in our class-agnostic setup below. The second loss term favors a time-consistent segmentation.

We denote the number of points, queries and pseudo ground-truth instances by $N_p$, $N_q$ and $N_o$, respectively. The loss is computed based on an affinity score $A_{ij}$\renaud{, which is the} \corentin{scalar product between the query-embeddings $j=1,\ldots,N_q$ and the point features $i=1,\ldots,N_p$, followed by a sigmoid.}
We also denote by $G \in [0, 1]^{N_p \times N_o}$ the  assignment matrix between the points and the objects as computed from  the pseudo-labels: $G_{io} = 1$ if point $i$ belongs to object $o$ according to the clustering, and $0$ otherwise.

\smallparagraph{Scan-wise training loss.} To compute the scan-wise loss, we need to find correspondences between queries and object instances. These correspondences are computed by minimizing a cost of assignment between them. The global cost matrix $C \in \mathbb{R}^{N_q \times N_o}$ between the queries and the object instances \renaud{is defined as:} 
\begin{align}
    C = \lambda_{\text{dice}} \, C^{\text{dice}} + \lambda_{\text{BCE}} \, C^{\text{BCE}} \> ,
\end{align}
where $C^{\text{dice}}$ is based on the Dice loss,
\begin{align}
    C^{\text{dice}}_{jo} = \frac{2 \; \sum_{i=1}^{N_p} {A}_{ij}{G}_{io}}{\sum_{i=1}^{N_p}{{A}_{ij}}^2 + \sum_{i=1}^{N_p}{{G}_{io}}^2} \> ,
\end{align}
and $C^{\text{BCE}}$ is based on the binary cross entropy,
\begin{fleqn} 
\begin{align}
    C^{\text{BCE}}_{jo} = \sum_{i=1}^{N_p} {G}_{io} \log({A}_{ij}) + (1-{G}_{io}) \log(1 - {A}_{ij}) \> .
\end{align}
\end{fleqn}
We use $\lambda_{\text{dice}} = 2$ and $\lambda_{\text{BCE}} = 5$, as in \cite{mask3d}. Unlike in \cite{mask3d}, we do not have any information about the semantic class of the objects in our unsupervised setting. We therefore do not use any classification loss related to semantic categories.
The actual matching between object instances $o$ and queries $j$ is obtained using the Kuhn–Munkres algorithm (aka Hungarian matching) with $C$ as cost matrix. Once this matching \renaud{$m(\cdot)$} is obtained, we minimize
\begin{align}
\sum_{o=1}^{\min (N_q, N_o)} \lambda_{\text{dice}} \, C^{\text{dice}}_{m(o)o} + \lambda_{\text{BCE}} \, C^{\text{BCE}}_{m(o)o}\> ,
\end{align}
where $m(o)$ is the index of the matched query for object $o$.
Finally, this process \renaud{is} 
repeated identically at $12$ different layers and scales of the ResUNet, and the final loss is 
a sum of all \renaud{corresponding loss computations,} 
as in~\cite{mask3d}.

\begin{figure*}[t]
\small
\centering
\def\svgwidth{\linewidth}
\input{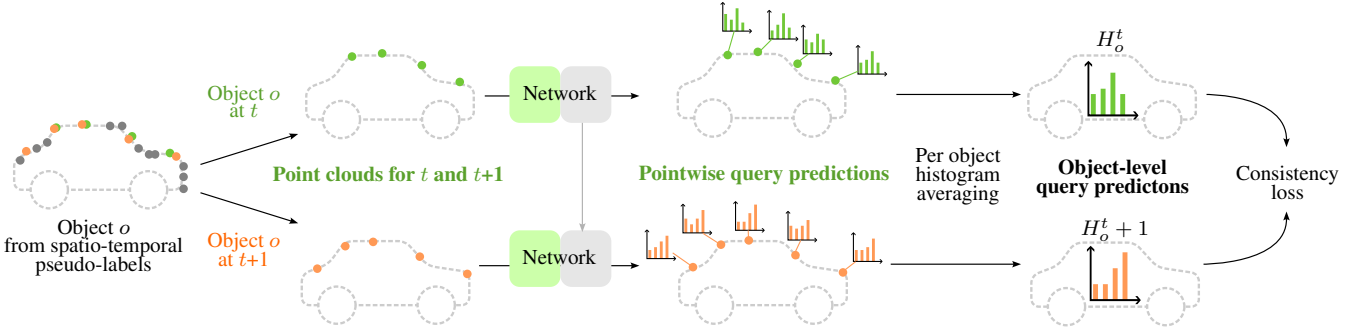}
\caption{\textbf{Computing the consistency loss}. Given an object, we compute the averages $H_{o}^{t}$ and $H_{o}^{t+1}$ of its pointwise query predictions over two time steps. The consistency loss is the cross-entropy between $H_{o}^{t}$ and $H_{o}^{t+1}$. 
}
\label{fig:consistency}
\end{figure*}

\smallparagraph{Time consistency loss.}
We aim at being able to segment and track object instances for a time period much longer than the time window seen during training. \renaud{In fact,} we can only train our network on two consecutive scans on our resources. To enforce time-coherent segmentation over \corentin{longer periods of} time, we propose the consistency loss described below.

Let us consider an object instance $o$ segmented at time~$t$ and $t$+1 with our pseudo-labeling strategy. For instance $o$~and \renaud{for} each query $q$, we compute the average similarity score (scalar product between query embedding and point features) with the points belonging to $o$ at time $t$ and $t$+1. We then transform these scores as a probability of assignment over the queries using a softmax layer (over the dimensions of the queries). We obtain two probability distributions for each object $o$: $H^t_o \in [0, 1]^{N_q}$ and $H^{t+1}_o \in [0, 1]^{N_q}$. 
%
\vincent{
To encourage temporal continuity, we minimize the relative entropy, i.e., the Kullback-Leibler divergence from ${H}_{o}^{t+1}$ to ${H}_{o}^{t}$
}
\begin{align}
\label{eq:consistency_loss}
\mathcal{L}^\text{cons}_o = D_{\text{KL}}({H}_{o}^{t} || {H}_{o}^{t+1}) \> .
\end{align}
\alex{We consider ${H}_{o}^{t}$ to be the target distribution and do not back-propagate through it (thanks to a stop-gradient operation). However, note that back-propagation goes through the two timesteps thanks to the auto-regressive architecture.}
\corentin{In practice, an equivalent gradient 
is obtained with a cross-entropy loss between ${H}_{o}^{t+1}$ and ${H}_{o}^{t}$, while keeping the stop-gradient operation on ${H}_{o}^{t}$.}

\smallparagraph{Global loss.} We \renaud{first} pretrain our network in a single scan setting, using the \renaud{scan-wise} loss function:
\begin{equation}
\frac{1}{N} \sum_{o=1}^{N} \left[ \lambda_{\text{dice}} \, C^{\text{dice}}_{m(o)o} + \lambda_{\text{BCE}} \, C^{\text{BCE}}_{m(o)o}\right] \> ,
\end{equation}
\corentin{with $N = \min (N_q, N_o)$.}
We then train on two consecutive scans by adding the time-consistency loss between the first and second scans:
\begin{equation}
\frac{1}{N} \sum_{o=1}^{N} \left[ \lambda_{\text{dice}} \, C^{\text{dice}}_{m(o)o} + \lambda_{\text{BCE}} \, C^{\text{BCE}}_{m(o)o} + \lambda_\text{cons} \mathcal{L}^\text{cons}_o \right] \> .
\end{equation}
In this second phase, costs $C^{\text{dice}}$ and $C^{\text{BCE}}$ are computed only on the second scan. Thus, the Hungarian matching used to compute  mapping $m(\cdot)$ uses only the query embedding at the second scan. We used $\lambda_\text{cons} = 1$ in our experiments.

\corentin{Further details about the architecture and training procedure are provided in appendix. 
}

\subsection{Query recycling mechanism}

\corentin{The number of queries in our architecture is fixed. \renaud{In our experiments, it is set} to $N_q = 300$, }
which off-the-shelf allows the segmentation and tracking of at most $300$ different instances over an entire sequence. 
\vincent{This is far lower than the number of objects in a sequence of thousands of scans. We need a process to handle the appearance of new objects and their disappearance \renaud{(when they go out of sight)} with a query recycling mechanism.}

\corentin{Our \renaud{recycling} process is \renaud{based on a} simple \renaud{heuristics}. After each scan at a given time $t$ is processed, we compute the barycenter of \renaud{the points associated to} every active query. Then for each active query $q$, we retrieve its previous barycenter, at $t_{\rm past}$. At this stage, we need a simple rule to determine if  query $q$ is still tracking the same object or tracking a new object. We do so by checking the distance between the two barycenters for $q$ at $t$ and $t_{\rm past}$. If this distance is smaller than $10$~m, then we assume the query is still tracking the same object; otherwise it is tracking a new object\renaud{, which was} never seen in the past.} \vincent{In this case, we assign a new object ID to the query.}

Note that the barycenters are computed in the coordinate system of the ego vehicle. We compute the distance between barycenters \emph{without} compensating for the motion of the ego-vehicle between time $t_{\rm past}$ and $t$, which means we do not need odometry information nor a side method to register the scans. Our criterion simply checks if, relative to the ego-vehicle, the segmented object appears in a completely different area than \renaud{where} it was before. This criterion is imperfect. \renaud{For instance,} a new object \renaud{could be} detected by the same query as an old object \renaud{if it were} appearing 
\renaud{at a location close to} the 
position \renaud{of} 
the old object relative to the ego-vehicle. \renaud{While this strategy still} 
works \renaud{reasonably} well in practice, \renaud{it leaves room for improvement.}

\section{Experiments}
\label{sec:experiments}

\subsection{Datasets}
\label{sec:datasets}

\smallparagraph{SemanticKITTI}\cite{semantickitti} is a Lidar dataset acquired in Germany, with a 64-beam Lidar capturing scans at 10 Hz. We use the usual split where sequences $00$ to $10$ but $08$ are used for (unsupervised) training, and sequence $08$ is used for validation. On the validation sequence, our metric are computed on \renaud{object} 
instances of \renaud{the following classes}: car, bicycle, motorcycle, truck, other-vehicle,
person, bicyclist, and motorcyclist.
\corentin{This dataset \renaud{contains} 
the longest sequences, with the validation sequence \renaud{consisting} 
of 4071 consecutive scans.}

\smallparagraph{PandaSet}\cite{pandaset} \corentin{contains data captured
with two different Lidar sensors, a rotating Pandar64 and a solid-state PandarGT\renaud{, at 10 Hz}. Following \cite{scalr}, we use GPS coordinates to create two spatially separated training and validation sets, using scans collected in San Francisco for training and the rest as the validation set. PandaSet does not contain instance ground truth. We create them by combining semantic segmentation and object detection labels in a process described in the supplementary material. In this work,
\gilles{we only experiment with the dense point clouds captured with PandarGT, as Pandar64 has similar characteristics to the Velodyne \renaud{HDL-64E} used in SemanticKITTI. We denote this set by Panda\renaud{Set-}GT.}}


\subsection{Baselines}
\label{sec:baselines}

We consider six different baselines. The first three baselines are 3DUIS \cite{3duis}, the unsupervised spatio-temporal segmentation used in TARL \cite{tarl} \renaud{(TARL-Seg)}, and our technique for generating pseudo-labels (dubbed \renaud{4D-Seg}). 
All of these baselines are able to segment and track over a limited time window. We thus propose an improved version of them by post-processing the results to find correspondences between instances obtain over two successive time windows. Note that, unlike \ours, none of these baselines are online, except 3DUIS, which does not provide spatio-temporal segments.

\smallparagraph{3DUIS}\cite{3duis} is, to the best of our knowledge, the only published method performing unsupervised instance segmentation on automotive Lidar point clouds. This method is working scan-wise and obtains instances by solving a graph-cut problem. \renaud{As solving} this graph-cut problem \renaud{already is} 
computationally intensive in the scan-wise setting, 
we do not apply graph-cuts directly in the spatio-temporal setting. Instead, we compare this method to ours in a scan-wise setting but also use the post-processing described below to turn the discovered 3D segments into 4D segments.

\smallparagraph{TARL's segments and our 4D segments}~are obtained using classical clustering~(HDBSCAN~\cite{hdbscan}) on scans accumulated after rigid registration (see \cref{sec:clustering}). 
Both methods provide spatio-temporal segments over a predefined time-window, unlike our method \ours which is online. Note as well that we do not need to register the scans in \ours. The clustering used in TARL was tuned for SemanticKITTI, with a time window corresponding to $12$ consecutive scans. Our 4D segments provides object instances over $40$ consecutive scans on \renaud{both} SemanticKITTI and PandaSet-GT.

\smallparagraph{Baseline improvement.} All of the above methods provide 4D segments over a fixed time window. By construction, they are unable to track an object over a temporal horizon longer than this time window, which impacts negatively their performance \renaud{relatively} 
to the metrics defined in \cref{sec:metrics}. 

We improve the performance of these baselines by matching instances discovered in two successive time-windows.
Given the last scan $\ell$ in a reference time window and the first scan $f$ in the next time window, we register these scans in the same coordinate system and compute the convex hull of all instances in $\ell$ and $f$. Whenever the convex hulls of two instances from different scans overlap with an IoU larger than $0.5$, the object instance in $f$ is assigned the object ID of the instance in $\ell$. Note that we use the ground-truth poses of the ego-vehicle only for these baselines---\ours does not need access to the vehicle poses.
\corentin{More details are provided in the supplementary material.}
We denote the improved baselines by 3DUIS++, TARL-Seg++, and 4D-Seg++.

\subsection{Metrics}
\label{metrics}
\label{sec:metrics}

\smallparagraph{Association score.} Our main metric is the association score used in \cite{aygun20214d} for 4D panoptic segmentation in Lidar data, and used in \cite{3duis} for unsupervised class-agnostic instance segmentation. We recall its definition below for completeness. 


Let $\mathcal{G}$ be the set of manually-annotated objects \renaud{(sets of points)} serving as ground truth for evaluation but unused during training, and $\mathcal{S}$ be the set of segments (objects) predicted by our network. Note that any object in $\mathcal{G}$ (similarly for $\mathcal{S}$) is a \emph{4D segments} containing the list of points belonging to that object at every timestep of the sequence. The temporal association score $\mathrm{S_{assoc}^{temp}}$ satisfies:
\begin{equation}
    \mathrm{S_{assoc}^{temp}} = \dfrac{1}{|\mathcal{G}|} \sum\limits_{g \in \mathcal{G}}\dfrac{1}{|g|} \sum\limits_{\substack{s \in S \\ s \cap g \neq 0}} \mathrm{TPA}(s,g) \mathrm{IoU}(s,g) \> ,
\end{equation}
where $\mathrm{TPA}\renaud{(s,g)} = |s \cap g|$ is the number of true positive associations between the 4D segments $s$ and $g$, and $\mathrm{IoU}$ is the intersection-over-union. 

A non-temporal version of the association score can be computed by considering that objects and predictions in different scans come from different instances. This version of the association metric, denoted $\mathrm{S_{assoc}}$, is the metric used in \cite{3duis} as this method works in a scan-wise setting. 

\smallparagraph{Best IoU.} In complement to the above score, we introduce another metric where, for each object in $\mathcal{G}$, we first find the best overlapping segments in $\mathcal{S}$ in terms of IoU, and then compute the average IoU with these best matches:
\begin{equation}
    \mathrm{IoU^*} = \dfrac{1}{|\mathcal{G}|} \sum\limits_{g \in \mathcal{G}} \max\limits_{s \in S} \mathrm{IoU}(s,g) \> .
\end{equation}
Note that this metric is computed using 4D segments: the higher the score, the better the temporal consistency. Note as well that, in this metric, a predicted segment $s$ can be a best match for two different ground-truth objects $g$ and $g'$~($g \neq g'$). It thus provides a measure of the best IoU we can reach with the available set predictions.

\begin{table}[t]
\small
\setlength{\tabcolsep}{4pt}
\centering
\begin{tabular}{lcccccc}
\toprule
\multirow{2}{*}{\fbox{Sem.KITTI}}
    & \multirow{3}{*}{\rotatebox[origin=c]{90}{\quad Online}}
    & \multicolumn{3}{c}{Unfiltered}
    & \multicolumn{2}{c}{Filtered}
\\
\cmidrule(lr){3-5}
\cmidrule(lr){6-7} 
    &
    & $\mathrm{S_{assoc}^{temp}}$     
    & $\mathrm{IoU^*}$
    & $\mathrm{S_{assoc}}$ 
    & $\mathrm{S_{assoc}^{temp}}$ 
    & $\mathrm{S_{assoc}}$ 
\\
\midrule
3DUIS w/o time     
    & \cmark
    & -
    & - 
    & 0.550
    & -
    & 0.768
\\
\rowcolor{blue!10}
\ours w/o time    
    & \cmark
    & - 
    & -
    & \textbf{0.715}
    & -
    & \textbf{0.811}
\\
\midrule
3DUIS++      
    & \cmark
    & 0.116
    & 0.214
    & 0.550
    & 0.148
    & 0.769
\\
TARL-Seg    
    & \xmark
    & 0.231
    & 0.353
    & 0.668
    & 0.264
    & 0.735
\\
TARL-Seg++     
    & \xmark
    & 0.317
    & 0.446
    & 0.617
    & 0.370
    & 0.678
\\
4D-Seg  
    & \xmark
    & 0.421
    & 0.529
    & 0.667
    & 0.486
    & 0.784
\\
4D-Seg++    
    & \xmark
    & 0.447
    & 0.513
    & 0.647
    & 0.512
    & 0.762
\\
\rowcolor{blue!10}
\ours     
    & \cmark
    & \textbf{0.482}
    & \textbf{0.568}
    & 0.696
    & \textbf{0.563}
    & 0.790
\\
\bottomrule
\end{tabular}
\caption{\textbf{Results on SemanticKITTI.} All scores are computed on the validation set of SemanticKITTI. The association scores are computed using the code of \cite{aygun20214d}, which, by default, is only applied on segments of more than $50$ points for any given scan; we report the corresponding scores with and without filtering.}
\label{table:results_sk}
\end{table}
\begin{table}[t]
\small
\setlength{\tabcolsep}{4pt}
\centering
\begin{tabular}{lcccccc}
\toprule 
 \multirow{2}{*}{\fbox{PandaSet-GT}}
    & \multirow{3}{*}{\rotatebox[origin=c]{90}{\quad Online}}
    & \multicolumn{3}{c}{Unfiltered}
    & \multicolumn{2}{c}{Filtered}
\\
\cmidrule(lr){3-5}
\cmidrule(lr){6-7}
    & 
    & $\mathrm{S_{assoc}^{temp}}$
    & $\mathrm{IoU^*}$
    & $\mathrm{S_{assoc}}$ 
    & $\mathrm{S_{assoc}^{temp}}$ 
    & $\mathrm{S_{assoc}}$ 
\\
\midrule
\rowcolor{blue!10}
\ours w/o time
    & \cmark
    & - 
    & - 
    & \textbf{0.562}
    & - 
    & 0.719
\\
\midrule
TARL-Seg
    & \xmark
    & 0.206
    & 0.286
    & 0.369
    & 0.390
    & 0.757
\\
4D-Seg
    & \xmark
    & \textbf{0.332}
    & \textbf{0.399}
    & 0.492
    & \textbf{0.503}
    & \textbf{0.740}
\\
\rowcolor{blue!10}
\ours   
    & \cmark
    & 0.209
    & 0.310
    & 0.531
    & 0.351
    & 0.688
\\
\bottomrule
\end{tabular}
\caption{\textbf{Results on Panda\renaud{Set-}GT} (validation set, see \cref{sec:datasets}). 
}
\label{table:results_pdgt}
\end{table}

\smallparagraph{Limitations.} The above metrics require 4D manual segments of each object. In existing datasets, such annotations are available only for \emph{thing} classes (cars, pedestrians, 
etc.). Thus, while our method is able to segment and track all objects (both \emph{stuff} and \emph{things}), these metrics \renaud{are only used to} measure 
performance 
on 
a subset of the objects.

\smallparagraph{Filtered metrics.} \corentin{The metrics 
used by \cite{3duis} and \cite{aygun20214d} include a scan-wise filter removing all 3D ground-truth segments with fewer than 50 points. We include these ``filtered'' metrics for comparability purposes. \gilles{We would like to recall nevertheless that this filtering removes small or distant objects from the metrics. We thus also compute the metrics without any filtering to provide a more complete assessment of the performance of the methods.} 
}

\subsection{Results of \ours}



\smallparagraph{On SemanticKITTI.}
We compare our method to all the considered baselines on SemanticKITTI. The results are presented in \cref{table:results_sk}. In the single-scan setting, \ours is better than 3DUIS. This shows the advantage of training end-to-end an instance segmenter rather than relying on a graph-cut method. In the online setting, we notice that post-processing the segments obtained by 3DUIS is not enough to compete with the segments obtained by TARL-Seg and 4D-Seg on both temporal metrics $\mathrm{S_{assoc}^{temp}}$ and $\mathrm{IoU^*}$. We also remark that 4D-Seg performs better than TARL-Seg. This is due to the longer temporal window used in 4D-Seg as both methods give similar scores on the scan-wise metric $\mathrm{S_{assoc}}$. We also observe that our post-processing over the segments of TARL-Seg and 4D-Seg improve the spatio-temporal quality of the segments. The best scores are obtained with our method.

\smallparagraph{On PandaSet-GT.}
\gilles{We notice in \cref{table:results_pdgt} that \ours surpasses TARL-Seg, except when filtering out all small ground-truth segments. Unlike on SemanticKITTI, \ours does not outperforme its pseudo-labels (4D-Seg). The large context window (4~s) for 4D-Seg with respect to sequence duration (8~s) on PandaSet-GT provides an advantage to 4D-Seg which has access to nearly the whole sequence for segmentating, while \ours works online. We also recall that \ours does not require precise registration of the scans, unlike the baselines. }

\subsection{Ablation study}
\begin{table}[t]
\renewcommand{\arraystretch}{0.8} 
\small
\setlength{\tabcolsep}{4pt}
\centering
\newcommand*\rotext[1]{\multicolumn{1}{R{30}{3.5em}}{\rlap{#1}\phantom{Scan-wise}}}
\begin{tabular}{@{}ccccccc@{}}
\toprule
\multicolumn{3}{c}{Training}
    & \multicolumn{2}{c}{Inference}
    & \multicolumn{2}{c}{Metric}
\\
\cmidrule(lr){1-3}
\cmidrule(lr){4-5}
\cmidrule(lr){6-7}
\rotext{Scan-wise}
    & \rotext{Autoreg.}
    & \rotext{$\mathcal{L}^\text{cons}_o$ \eqref{eq:consistency_loss}}
    & \rotext{Scan-wise}
    & \rotext{Autoreg.}
    & \rotext{$\mathrm{S_{assoc}^{temp}}$}
    & \rotext{$\mathrm{S_{assoc}}$}

\\
\midrule
\multicolumn{5}{c}{\cellcolor{gray!20}{4D-Seg}}
    & 0.420
    & 0.667
\\
\midrule
\cmark
    & -           
    & \xmark
    & \cmark
    & -
    & -
    & 0.715
\\
\cmark 
    & -           
    & \xmark
    & -
    & \cmark
    & 0.330
    & 0.667
\\
- 
    & \cmark           
    & \xmark
    & -
    & \cmark     
    & 0.446
    & 0.727
\\
- 
    & \cmark           
    & \cmark
    & -
    & \cmark 
    & 0.482
    & 0.696
\\
\bottomrule
\end{tabular}
\caption{\textbf{Ablation study 
on the validation set of SemanticKITTI.} Influence of the training and inference procedure (scan-wise or auto-regessive) as well as the effect of the loss \eqref{eq:consistency_loss} on the performance.}
\label{table:ablation_sk}
\end{table}

We conduct an ablation study on the components of \ours on SemanticKITTI. The results are reported in \cref{table:ablation_sk}.

First, we remark that, surprisingly, training our network in a scan-wise setting but doing the inference in an auto-regressive manner (second row of the table) already provides reasonable segmentation of instances in 4D, with scores not so far from \renaud{those of} 4D-Seg. This indicates that using as input the query embedding of the past scan already allows to track some object instances without explicit supervision. This justifies the auto-regressive use of the queries. 

Second, as seen from the third row of \cref{table:ablation_sk}, training the network in an auto-regressive manner permits \renaud{us} to surpass the 4D-Seg baseline 
(our pseudo-labels). Note that, in this case, there is no explicit constraint or regularization that favors the discovery of spatially and temporally consistent instances. This behavior is only achieved implicitly via the auto-regressive use of the query embeddings and the use of spatio-temporal pseudo-labels. Yet, these two ingredients are enough to surpass 4D-Seg.

Finally, adding the time consistency loss further boosts the performance on the spatio-temporal metrics. 


\subsection{Case of low density point clouds}

\begin{table}[t]
\small
\setlength{\tabcolsep}{4pt}
\centering
\begin{tabular}{lcccccc}
\toprule 
 \multirow{2}{*}{\fbox{nuScenes}}
    & \multirow{3}{*}{\rotatebox[origin=c]{90}{\quad Online}}
    & \multicolumn{3}{c}{Unfiltered}
    & \multicolumn{2}{c}{Filtered}
\\
\cmidrule(lr){3-5}
\cmidrule(lr){6-7}   
    &
    & $\mathrm{S_{assoc}^{temp}}$
    & $\mathrm{IoU^*}$
    & $\mathrm{S_{assoc}}$ 
    & $\mathrm{S_{assoc}^{temp}}$ 
    & $\mathrm{S_{assoc}}$ 
\\
\midrule
\rowcolor{blue!10}
\ours w/o time
    & \cmark
    & - 
    & - 
    & \textbf{0.390}
    & - 
    & \textbf{0.570}
\\
\midrule
TARL-Seg
    & \xmark
    & 0.085
    & 0.156
    & 0.189
    & 0.253
    & 0.475
\\
4D-Seg
    & \xmark
    & \textbf{0.163}
    & \textbf{0.246}    
    & 0.287
    & \textbf{0.350}
    & 0.546
\\
\rowcolor{blue!10}
\ours   
    & \cmark
    & 0.126
    & 0.221
    & 0.356
    & 0.270
    & 0.561
\\
\bottomrule
\end{tabular} \\
\caption{\textbf{Results on nuScenes} (validation set).
}
\label{table:results_ns}
\end{table}

\begin{table}
\centering
\scriptsize
\setlength{\tabcolsep}{3.2pt}
\begin{tabular}{llc|l|l|l|l}
\toprule
&      & & \multicolumn{4}{c}{Percentage of training data}  \\
&    Pre-training & Inputs & \multicolumn{1}{c|}{1\%} & \multicolumn{1}{c|}{10\%} & \multicolumn{1}{c|}{50\%} & \multicolumn{1}{c}{100\%}\\
    \midrule
    \midrule
\rowcolor{black!10}
\cellcolor{white}&No pre-training                      & xyz + i & 
35.0 & 57.3 & 69.0 & 71.2 \\
\rowcolor{black!10}
\cellcolor{white}&PointContrast~\cite{pointcontrast}   & xyz + i & 
37.0 \textcolor{MyGreen}{+2.0}& 58.9 \textcolor{MyGreen}{+1.6} & 69.4 \textcolor{MyGreen}{+0.4} & 71.1 \textcolor{MyRed}{-0.1} \\
\rowcolor{black!10}
\cellcolor{white}&ALSO~\cite{also}                     & xyz + i & 
37.4  \textcolor{MyGreen}{+2.4} & \bf 59.0 \textcolor{MyGreen}{+2.7} & 69.8  \textcolor{MyGreen}{+0.8} & 71.8  \textcolor{MyGreen}{+0.6}\\
\rowcolor{black!10}
\cellcolor{white}&BEVContrast~\cite{bevcontrast}       & xyz + i & 
\bf 37.9  \textcolor{MyGreen}{+2.9} & \bf 59.0 \textcolor{MyGreen}{+2.7} & \bf 70.5 \textcolor{MyGreen}{+1.5} & \bf 72.2 \textcolor{MyGreen}{+1.0}\\
\cmidrule{2-7}
&No pretraining        & depth + i & 
35.4 & 60.2 & 71.3 & 71.3 \\
&\ours w/o time  & depth + i & 
35.3 \textcolor{MyRed}{-0.1} & 60.9 \textcolor{MyGreen}{+0.7} & \bf 72.2 \textcolor{MyGreen}{+0.9} & \bf 74.0 \textcolor{MyGreen}{+2.7}\\
\rowcolor{blue!10}
\multirow{-7}{*}{\rotatebox[origin=c]{90}{nuScenes}} 
\cellcolor{white}& \ours     & depth + i & 
35.4 & \bf 61.9 \textcolor{MyGreen}{+1.7} & 72.0 \textcolor{MyGreen}{+0.7} & 73.6 \textcolor{MyGreen}{+2.3}\\
    \midrule
    \midrule
\rowcolor{black!10}
\cellcolor{white}&No pre-training                      & xyz + i & 46.2 & 57.6  & 61.8  & 62.7  \\
\rowcolor{black!10}
\cellcolor{white}&PointContrast~\cite{pointcontrast}   & xyz + i & 47.9 \textcolor{MyGreen}{+1.7} & 59.7 \textcolor{MyGreen}{+2.1} & 62.7 \textcolor{MyGreen}{+0.9} & 63.4 \textcolor{MyGreen}{+0.7} \\
\rowcolor{black!10}
\cellcolor{white}&SegContrast~\cite{segcontrast}       & xyz + i & 48.9 \textcolor{MyGreen}{+2.7} & 58.7 \textcolor{MyGreen}{+1.1} & 62.1 \textcolor{MyGreen}{+0.3} & 62.3 \textcolor{MyRed}{-0.4} \\
\rowcolor{black!10}
\cellcolor{white}&STSSL~\cite{stssl}                   & xyz + i & 49.4 \textcolor{MyGreen}{+3.2} & 60.0 \textcolor{MyGreen}{+2.4} & 62.9 \textcolor{MyGreen}{+1.1} & 63.3 \textcolor{MyGreen}{+0.6} \\
\rowcolor{black!10}
\cellcolor{white}&ALSO~\cite{also}                     & xyz + i & 50.0 \textcolor{MyGreen}{+3.8} & 60.5 \textcolor{MyGreen}{+2.9} & \bf 63.4 \textcolor{MyGreen}{+1.6} & 63.6 \textcolor{MyGreen}{+0.9} \\
\rowcolor{black!10}
\cellcolor{white}&TARL~\cite{tarl}                     & xyz + i & 52.5 \textcolor{MyGreen}{+6.3} & 61.2 \textcolor{MyGreen}{+3.6} & \bf 63.4 \textcolor{MyGreen}{+1.6} & 63.7 \textcolor{MyGreen}{+1.0} \\
\rowcolor{black!10}
\cellcolor{white}&BEVContrast~\cite{bevcontrast}       & xyz + i & \bf 53.8 \textcolor{MyGreen}{+7.6} & \bf 61.4 \textcolor{MyGreen}{+3.8} & \bf 63.4 \textcolor{MyGreen}{+1.6} & \bf 64.1 \textcolor{MyGreen}{+1.4} \\
\cmidrule{2-7}
&No pretraining        & depth + i & 46.6  & 59.8 & 62.2 & 63.4 \\
&\ours w/o time  & depth + i & 49.0 \textcolor{MyGreen}{+2.4}    & 61.8 \textcolor{MyGreen}{+2.0} & \bf64.9 \textcolor{MyGreen}{+2.7} & 65.1 \textcolor{MyGreen}{+1.7}\\
\rowcolor{blue!10}
\multirow{-10}{*}{\rotatebox[origin=c]{90}{SemanticKITTI}} 
\cellcolor{white}&
\ours     & depth + i & \bf50.2 \textcolor{MyGreen}{+3.6} & \bf62.4 \textcolor{MyGreen}{+2.6} & \bf64.9 \textcolor{MyGreen}{+2.7} & \bf65.9 \textcolor{MyGreen}{+2.5}\\
\bottomrule
\end{tabular}
\caption{\corentin{\textbf{Fine-tuning for semantic segmentation.} We show the quality \renaud{(mIoU\% after fine-tuning)} of \renaud{the features we learn} 
self-supervisedly. Other methods, in grey, have a similar training setup but a different network architecture and inputs. \renaud{`i' is for intensity.}}}
\label{table:bb-ft-together}
\end{table}

Our method, like \cite{tarl,segcontrast,3duis}, relies on the extraction of accurate segments for pseudo-labeling. Such a segmentation is far easier when starting from dense point clouds, as available in SemanticKITTI and PandaSet-GT, than when starting from low-density point clouds. This probably explains why related methods do not conduct experiments on datasets like nuScenes, where point clouds are much sparser than in SemanticKITTI. To test the limits of our method and baselines on low-density point clouds, we apply them on nuScenes \cite{nuscenes} (see description in Appendix).

As on the other datasets, we notice (cf.\ \cref{table:results_ns}) that 4D-Seg surpasses TARL-Seg thanks to its longer temporal window. \ours performs better than 4D-Seg on the non-temporal metric $\mathrm{S_{assoc}}$ but remains below on the temporal metrics: \ours does not surpass its pseudo-labels. We recall nevertheless that \ours work online and without accumulation and registration of scans, unlike 4D-Seg. Let us also mention that we noticed a slight drop of performance for \ours when using the consistency loss on nuScenes. We hypothesize that the small point density leads to worse pseudo-labels for small or distant objects, which in turn creates more noise on the segments used in the consistency loss.

\subsection{Fine-tuning the backbone for self-supervision}

As an additional experiment, we study the potential of \ours as a self-supervised method for semantic segmentation. We use the sparse U-Net~\cite{mask3d} backbone pre-trained by \ours as a point-based segmenter.
Following the protocol of \cite{also,bevcontrast} on nuScenes and SemanticKITTI, we finetune the pretrained backbone with different amount of annotated data (1\%, 10\%, 50\% and 100\%).
The results are presented in \cref{table:bb-ft-together}.
For comparison, we also report the scores from the literature~\cite{bevcontrast}. \gilles{Let us highlight nevertheless that these methods use different variants of sparse U-Net, with different input features. Yet all sparse U-Nets attain comparable baseline results when using 100\% of labeled data and no pretraining.} 

\gilles{We notice that our pretrained sparse backbone underperforms at very low data regime, but reaches better performance than state-of-the-art self-supervised techniques at high annotation percentages.}
\alex{\renaud{It could be due to} 
the size of the self-training head, i.e., the layers on top of the sparse U-Net.
Indeed our head is a complete multi-scale transformer, as opposed to previous methods that are at most 3-layered MLPs.
In the latter case, the output feature maps are more likely to be linearly separable, thus requiring less finetuning 
and performing better in low-data regime.
A similar behavior is observed for MAE~\cite{he2022masked} where the pretraining head is an 8-block transformer and the best performance are reached for very long finetuning.} We also observe that going from single-scan to multi-scan models improves the performance. 

\section{Conclusion}
We showed that we could segment and track objects in Lidar data without requiring any manual annotations. This is made thanks to an offline method generating pseudo-labels, and to a novel auto-regressive architecture that performs robustly online once trained on these noisy pseudo-labels only. We hope that our approach will inspire other authors to consider other tracking problems where annotations are not available. 

\paragraph{Acknowledgements:}
This work was performed using HPC resources from GENCI–IDRIS (Grants 2023-AD011013765R1).
This work was supported in part by the French Agence Nationale de la Recherche (ANR) grant MultiTrans (ANR21-CE23-0032).

{
    \small
    \bibliographystyle{ieeenat_fullname}
    \bibliography{main}
}

\clearpage
\maketitlesupplementary
\appendix







In \cref{sec:pseudolab}, we detail the construction of pseudo-labels. In \cref{sec:unitdetails}, we provide details on the architecture and training of \ours. In \cref{sec:baselinedetails}, we provide details on our improvement of the baselines. Finally, in \cref{sec:experim}, we detail the experiments on low density point clouds and provides more qualitative results (visualizations).

\section{Pseudo-label construction}
\label{sec:pseudolab}

In this section, we describe the practical implementation details of the construction of pseudo-labels: for our 4D-Seg~(\cref{sec:4dsegdetails}), for TARL-Seg (\cref{sec:tarlsegdetails}) regarding other datasets than the one used in \cite{tarl}, and for the case of PandaSet (\cref{sec:pdsetdetails}), which does not contain ground-truth instance information.

\subsection{4D-Seg implementation details}
\label{sec:4dsegdetails}

\paragraph{Ground point segmentation.} 

To construct our 4D segments, we first apply Patchwork~\cite{patchwork} or Patchwork++~\cite{patchworkpp} on each individual scan to segment ground points.

For SemanticKITTI, we used Patchwork with the set of parameters used by TARL~\cite{tarl}. For nuScenes, which uses a 32-beam Lidar, we used Patchwork++ with a sensor height of $1.840$, thresholds for seeds and distance respectively of $0.5$ and $0.25$, and a minimum range of $2$. PandaSet being provided in world coordinates, we register each Lidar scan to its local reference frame before applying Patchwork++ with default parameters.
The respective quality of Patchwork++ for nuScenes and PandaSet-GT varies little with the parameters.

\paragraph{Temporal aggregation.} 

Points are then aggregated on a common reference frame, along with their temporal index~(an integer), 
for $40$ consecutive scans on SemanticKITTI and PandaSet-GT, and $80$ on the less dense nuScenes dataset, that is however captured at $20$\,Hz while both SemanticKITTI and PandaSet-GT are scanned at 10\,Hz.

A grid sampling with a voxel size of $5$\,cm along the spatial axis and $5$ time steps along the temporal axis is then applied to reduce the computation burden. While this grid sampling may reduce the overall quality of the segmentation, it considerably speeds up clustering, enabling longer temporal windows compared to TARL~\cite{tarl}.

\paragraph{Clustering.} 

Points that have little chances to belong to an object instance are put aside before clustering.  It concerns estimated ground points as well as points extremely close to the sensor, which are likely to be outliers or part of the roof of the ego-vehicle.

For the HDBSCAN spatio-temporal clustering, the temporal dimension is multiplied by $0.03$ to reduce its importance relative to the spatial dimensions. On nuScenes, the $z$ coordinate is also multiplied by $0.25$ to alleviate a splicing issue occurring in HDBSCAN due to the large vertical distance between Lidar beams of the Velodyne HDL-32E sensor. HDBSCAN's minimum number of samples and minimum cluster size are respectively set to $1$ and $300$ to obtain a large set of clusters, with very few discarded points. 

\paragraph{Instance ID assignment.} 

Each resulting segment is assumed to correspond to a different object instance and is given a separate index.  Ground points and discarded points are given special instance indexes. Points culled by the grid sampling process are added back, with the same index as non-culled points in their voxel.

\paragraph{Processing and output.} 

This segmentation is computed a single time, as a pre-processing for our method, and provides temporal windows of consecutive scans with consistent segments, sharing IDs between scans, and special ground and unknown segments.

\subsection{Implementing TARL-Seg on more datasets}
\label{sec:tarlsegdetails}

TARL \cite{tarl} was implemented by its authors on Semantic\-KITTI solely. For comparison purposes, we reimplemented TARL-Seg for PandaSet-GT and nuScenes.

We used Patchwork++~\cite{patchwork} for ground point segmentation as it requires less tuning that Patchwork~\cite{patchwork}, and has been tested with success on 4D-Seg (see \cref{sec:4dsegdetails}). 

As in the original TARL-Seg segment construction for SemanticKITTI, we do not apply any grid sampling. We thus use context windows of 12 time steps for PandaSet-GT and 24 for nuScenes.

Unlike in 4D-Seg, we do not use time as extra information for clustering.
However, on nuScenes, we apply the splicing trick described previously (see \cref{sec:4dsegdetails}),\vincentrmk{in A1.clustering?} as otherwise results are unfairly poor. The HDBSCAN parameters are the same as those used in TARL for SemanticKITTI.

\subsection{Instance ground-truth on PandaSet-GT}
\label{sec:pdsetdetails}

The PandaSet dataset does not come with panoptic labels. However, it comes with semantically-labeled 3D bounding boxes, as well as semantic segmentation for all Lidar points, with matching semantic classes between both modalities. 

To obtain instance labels, we look for all points inside a box that share the same class as the box; we then assign to these points an instance ID based on their associated box. We then relax slightly this point assignment by also incorporating into instance segments neighboring points of the same class, on the condition that their distance to the closest point of the instance is less than $1$\,m. In case a point is less than $1$\,m away from several instances, it is assigned to the closest one.

The code to generate these instance labels for PandaSet-GT will be supplied with the code for \ours.

\section{UNIT implementation details}
\label{sec:unitdetails}

\subsection{Architecture}
\label{sec:archidetails}

Our architecture is largely inherited from \cite{mask3d}. The backbone is a MinkUnet34~\cite{minkowski} with a voxel size of $15$\,cm. To save some computation resources, we use mixed precision using bfloat16, as well as flash attention~\cite{flashattention}. 

We apply two random augmentations during scan-wise training: a random scaling of $\pm 10\%$ and a random rotation 0-360°. No augmentation is applied when training in a temporal setup.

The input features are the Euclidean distance to the sensor and the returned Lidar intensity.


\subsection{Training protocol}
\label{sec:traningdetails}

In a first phase, we pretrain our network in a single-scan setting. The training is done in mixed precision using AdamW~\cite{adamw}, a cosine annealing scheduler for the learning rate with initial value of $10^{-4}$, a weight decay of $10^{-2}$, and a batch size of $3$. Due to differences in dataset sizes, we pretrain the network for $50$ epochs on SemanticKITTI, $150$ on PandaSet-GT and $4$ epochs on nuScenes, which corresponds to roughly $300$k training iterations for SemanticKITTI and nuScenes, and $200$k iterations for PandaSet-GT. 

In the second phase, we finetune the network using pairs of consecutive scans as input. We use the same hyperparameters for the optimizer. We use the same number of epochs,
and a batch size of $4$ for SemanticKITTI and $5$ for other methods, which corresponds to roughly $200$k training iterations for SemanticKITTI and nuScenes, and $100$k training iterations for PandaSet-GT.

\section{Baseline improvement details}
\label{sec:baselinedetails}

All the baselines methods that we initially considered either operate on single scans (3DUIS) or are consistent over a limited time window (TARL-Seg, 4D-Seg). We extended those baselines to stitch objects between two scans with separate ID predictions. For 3DUIS, we stich all pairs of successive scans; for TARL-Seg and 4D-Seg, we stich every pair of non-overlapping successive windows. This results in our improved versions of the baselines (3DUIS++, TARL-Seg++, 4D-Seg++).

The process is as follows. Given the last scan $\ell$ and the new scan $f$, registered in a common reference frame, we compute the convex hulls of all predicted instances in $f$ and in $\ell$. For each hull in $f$, we compute the IoU with all hulls in $\ell$ using a Monte Carlo method, and associate the two instances if the IoU is greater than 0.5. 

This process is extremely slow, and to speed it up, we decimate the instances to at most 200 points before the computation of the convex hull, estimate the IoU using 1000 samples, and stop the computation at the first hull found greater than $0.5$. An interesting observation is that this criterion is not strictly unique, meaning that multiple instances in $f$ can be associated to a single instance in $\ell$. This alone explains why those improved baselines performs usually poorer in scan-wise metrics.

\section{Experiments}
\label{sec:experim}

\subsection{Datasets}

SemanticKITTI~\cite{semantickitti} and PandaSet-GT~\cite{pandaset} are described in the main paper. We give here an overview of nuScenes~\cite{nuscenes}.

nuScenes is a Lidar dataset acquired in Boston and Singapore, with a 32-beam Lidar capturing scans at 20\,Hz. We use the official train/val split with 700 sequences for (unsupervised) training and 150 sequences for validation. We leverage all the lidar scans acquired at 20\,Hz during training and inference, for all methods. However, all metrics are computed on a subset of the validation scans as the point-wise annotations are provided at 2\,Hz only. For evaluation, we use the official panoptic annotations provided for the instances of: barrier, bicycle, bus, car, construction vehicle, motorcycle, pedestrian, traffic cone, trailer and truck.

\subsection{Qualitative results: more visualizations}

For qualitative inspection, we provide in the following figures more visualizations of the segments of 3DUIS++, TARL-Seg 4D-Seg, \ours and the ground truth, both on SemanticKITTI and PandaSet-GT.

\begin{figure*}[p]
    \centering 
    \begin{minipage}{0.48\linewidth}
    \includegraphics[width=\linewidth]{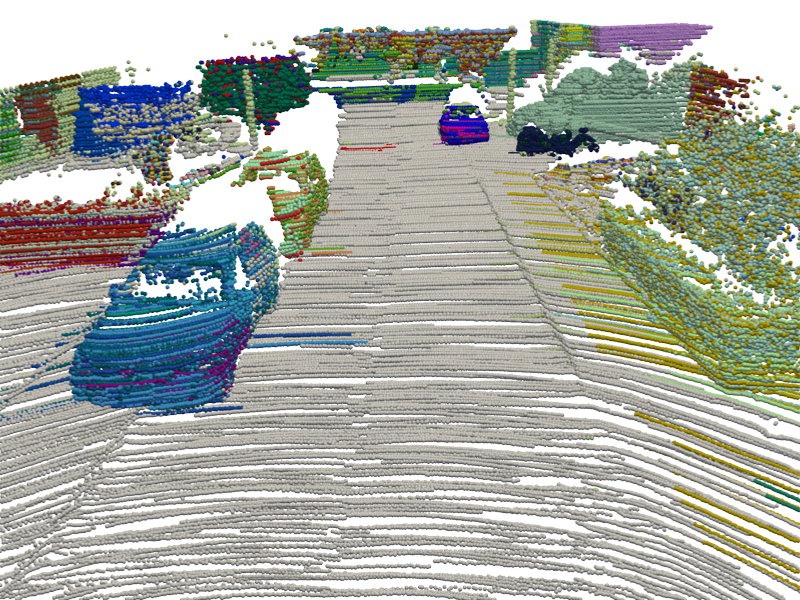}
    \centering \small 3DUIS++: offline stitching (`++') of online 3DUIS \\ single-scan segments
    \end{minipage}~~~
    \begin{minipage}{0.48\linewidth}
    \includegraphics[width=\linewidth]{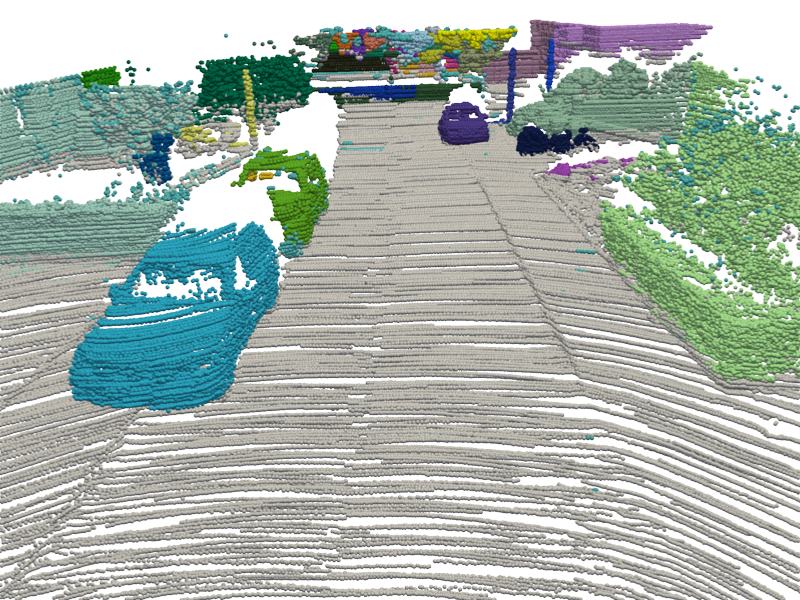}
    \centering \small 4D-Seg: our offline 4D segments computed on \\ aggregated scans
    \end{minipage}
    \\
    \begin{minipage}{0.48\linewidth}
    \includegraphics[width=\linewidth]{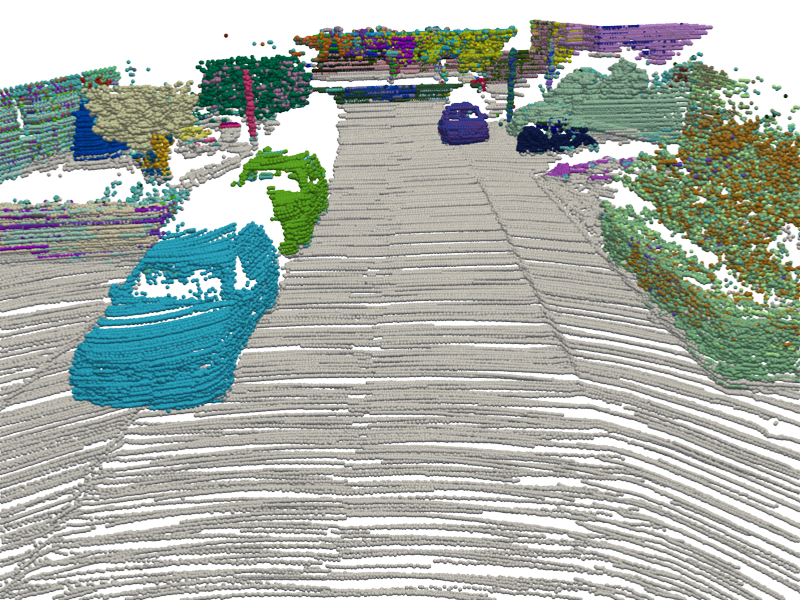}
    \centering \small \ours (trained on 4D-Seg): overlay of our successive \\ online single-scan segments 
    \end{minipage}~~~
    \begin{minipage}{0.48\linewidth}
    \includegraphics[width=\linewidth]{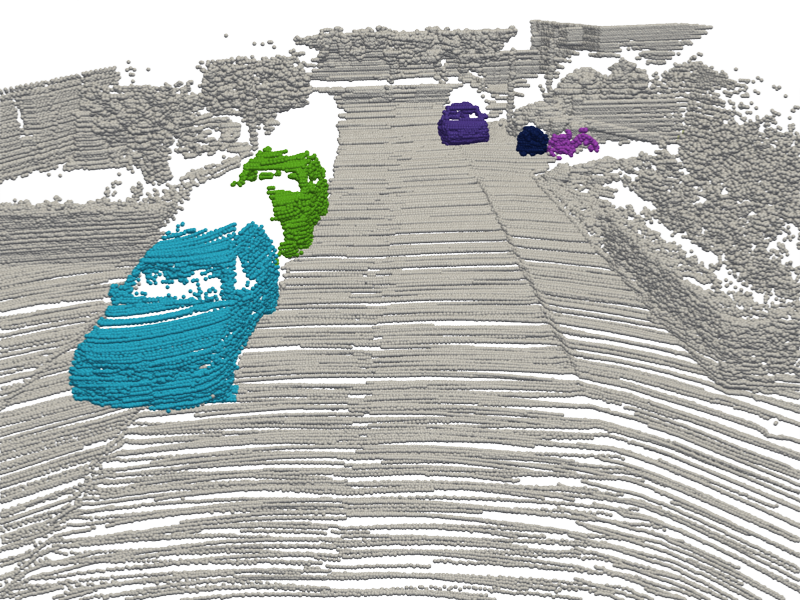}
    \centering \small GT: overlay of ground-truth single-scan segments \\ $\vphantom{\text{Pg}}$
    \end{minipage}
    \caption{Visualization of object instances across time, obtained on a sample scene of SemanticKITTI. Different instances are assigned colors that are random but (tentatively) consistent over time, forming segments in offline aggregated scans or overlaid online scans. 
    \newline\mbox{}\hspace*{12pt} 
    In this sample, which features a static scene, 4D-Seg is significantly cleaner than 3DUIS++ where single objects, such as the foreground car, have points belonging to several instances and leaking into the ground. \ours is online (it inputs and outputs one scan at a time) while 4D-Seg is offline (operating on an aggregation of scans), and is here slightly better than 4D-Seg on the cars, but not as good on ``stuff'' such as vegetation or walls.
    \newline\mbox{}\hspace*{12pt}
    Please note that we (4D-Seg and \ours) obtain more labels than in the ground truth as our class-agnostic segmentation include all objects and stuff, such as trees or buildings, while the ground truth is restricted to a few selected object classes.}
\end{figure*}

\begin{figure*}[p]
    \centering 
    \begin{minipage}{0.48\linewidth}
    \includegraphics[width=\linewidth]{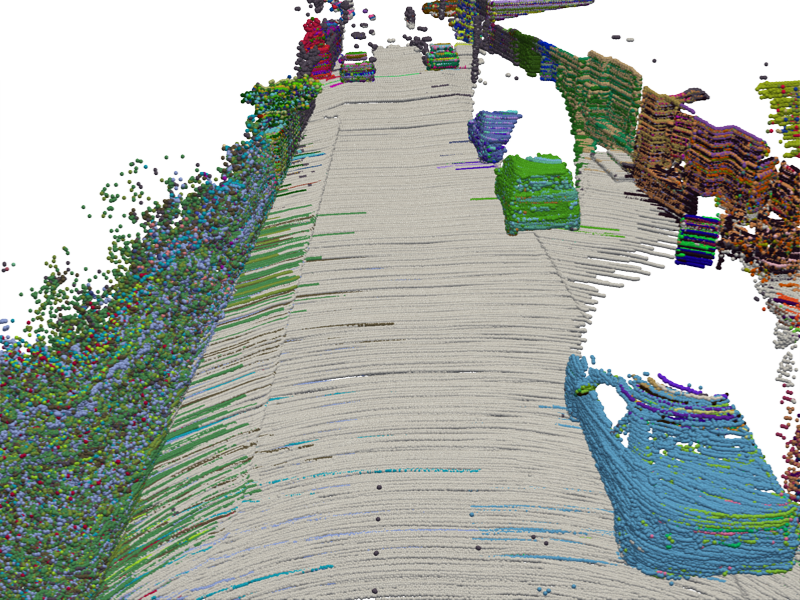}
    \centering \small 3DUIS++: offline stitching (`++') of online 3DUIS \\ single-scan segments
    \end{minipage}~~~
    \begin{minipage}{0.48\linewidth}
    \includegraphics[width=\linewidth]{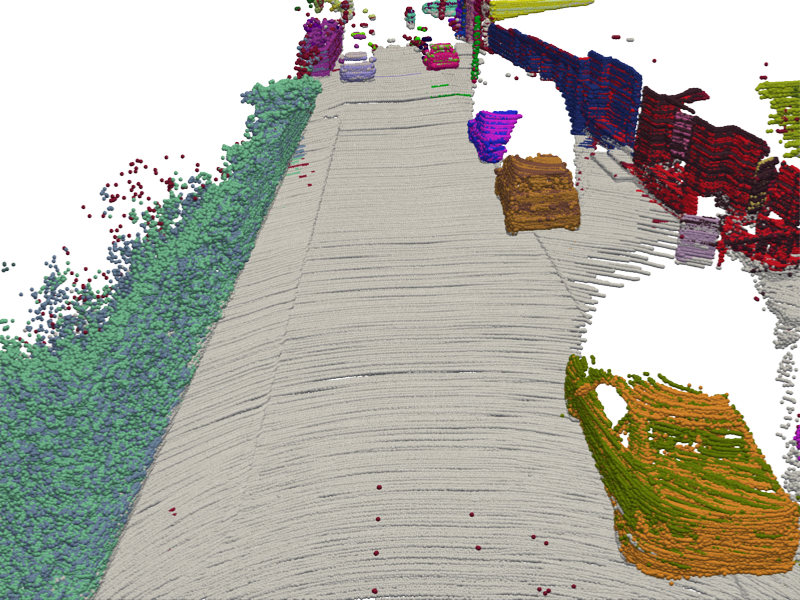}
    \centering \small 4D-Seg: our offline 4D segments computed on \\ aggregated scans
    \end{minipage}
    \\ $\vphantom{x}$ \\
    \begin{minipage}{0.48\linewidth}
    \includegraphics[width=\linewidth]{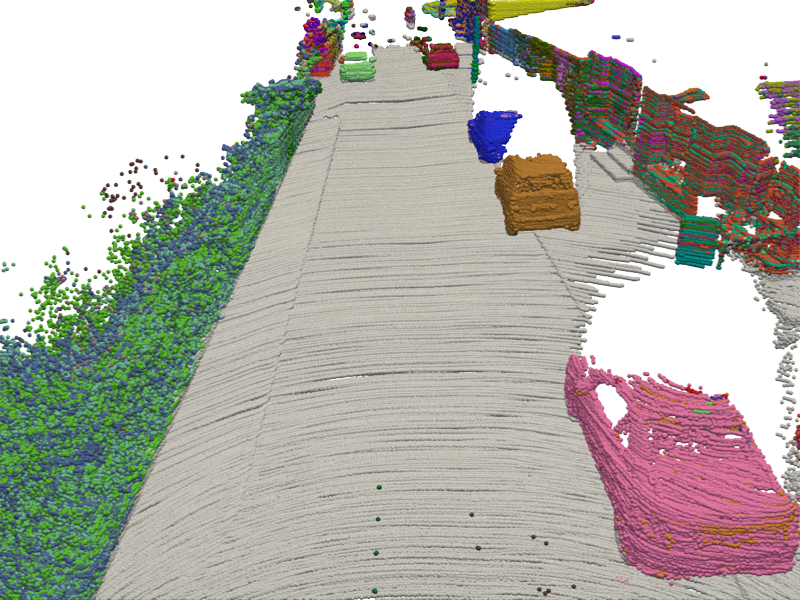}
    \centering \small \ours (trained on 4D-Seg): overlay of our successive \\ online single-scan segments
    \end{minipage}~~~
    \begin{minipage}{0.48\linewidth}
    \includegraphics[width=\linewidth]{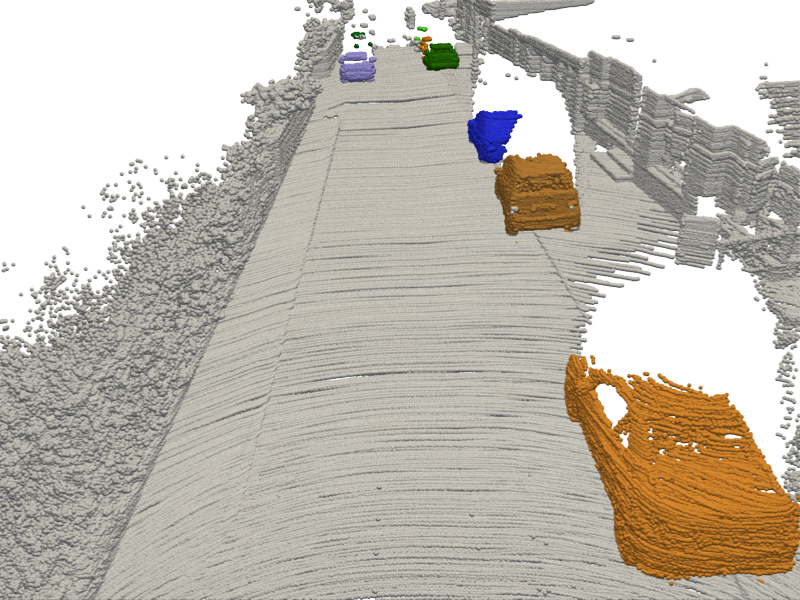}
    \centering \small GT: overlay of ground-truth single-scan segments \\ $\vphantom{\text{Pg}}$
    \end{minipage}
    \caption{Visualization of object instances across time, obtained on a sample scene of SemanticKITTI. Different instances are assigned colors that are random but (tentatively) consistent over time, forming segments in offline aggregated scans or overlaid online scans.
    \newline\mbox{}\hspace*{12pt} 
    In this sample, which features a static scene, 4D-Seg is not perfect but still better than 3DUIS++: there is no object segment leaking onto the ground, and the objects are assigned a smaller number of different IDs. For instance, the foreground car in 3DUIS++ is made of 6 different IDs while only 2 are used in 4D-Seg. It is all the more so when the objects are further away, e.g., concerning the cars at the end of the street. The training of \ours on 4D-Seg, while it is harder because it works online on single scans, rather than offline on aggregated scans as does 4D-Seg, regularizes the segmentation. For instance, the foreground car with \ours is ``now'' almost exclusively made of a single ID. Even further away objects benefit from that regularization, e.g., the cars at the end of the street. The instance ground truth shown here was obtained in a process detailed in \cref{sec:pdsetdetails}
    \newline\mbox{}\hspace*{12pt}
    Please note that we (4D-Seg and \ours) obtain more labels than in the ground truth as our class-agnostic segmentation include all objects and stuff, such as trees or buildings, while the ground truth is restricted to a few selected object classes.}
\end{figure*}

\begin{figure*}[p]
    \centering 
    \begin{minipage}{0.48\linewidth}
    \includegraphics[width=\linewidth]{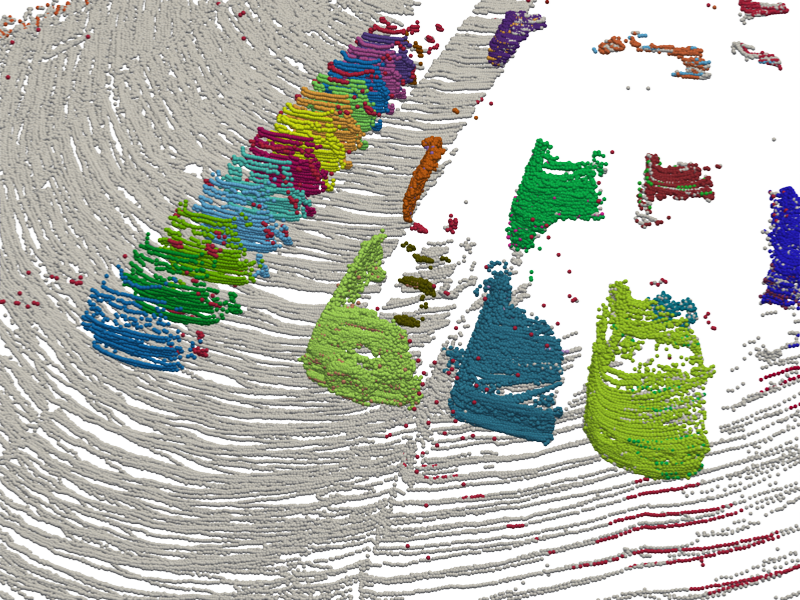}
    \centering \small TARL-Seg: offline 4D segments from TARL computed on \\ aggregated scans
    \end{minipage}~~~
    \begin{minipage}{0.48\linewidth}
    \includegraphics[width=\linewidth]{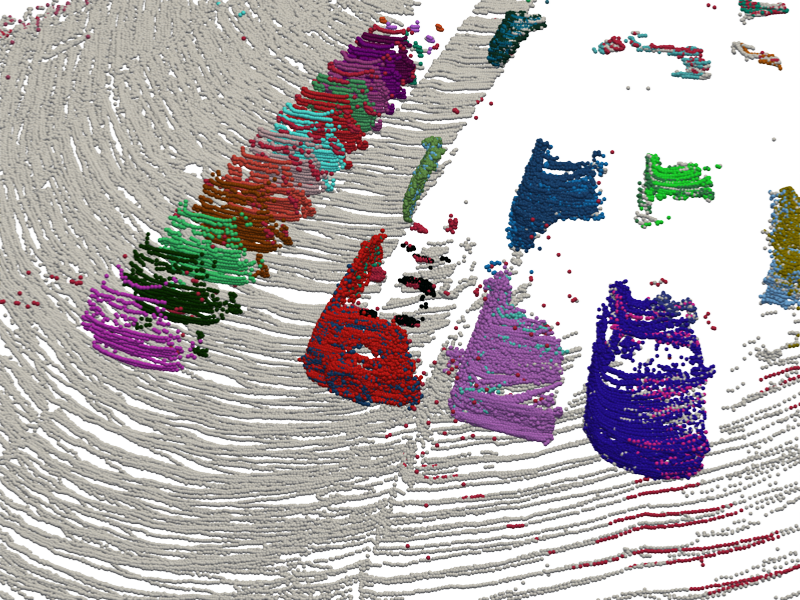}
    \centering \small 4D-Seg: our offline 4D segments computed on \\ aggregated scans
    \end{minipage}
    \\ $\vphantom{x}$ \\
    \begin{minipage}{0.48\linewidth}
    \includegraphics[width=\linewidth]{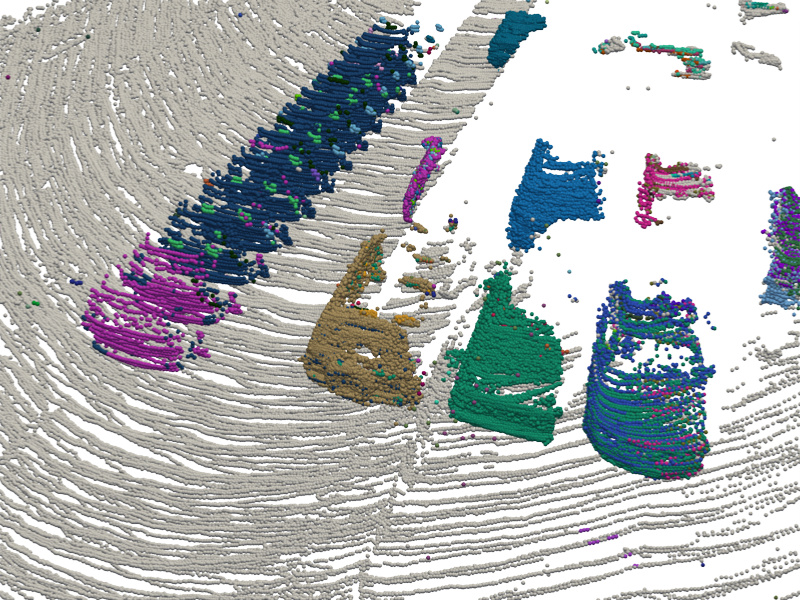}
    \centering \small \ours (trained on 4D-Seg): overlay of our successive \\ online single-scan segments
    \end{minipage}~~~
    \begin{minipage}{0.48\linewidth}
    \includegraphics[width=\linewidth]{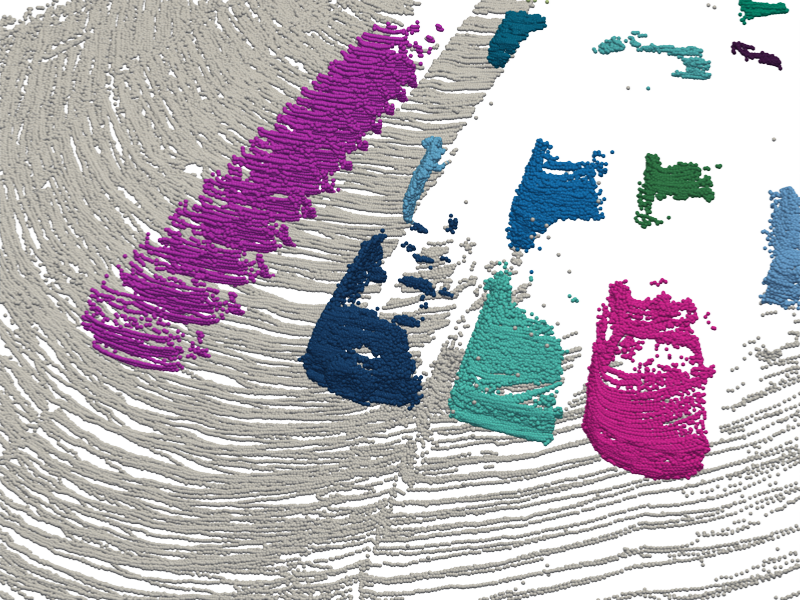}
    \centering \small GT: overlay of ground-truth single-scan segments \\ $\vphantom{\text{Pg}}$
    \end{minipage}
    \caption{Visualization of object instances across time, obtained on a sample scene of PandaSet-GT. Different instances are assigned colors that are random but (tentatively) consistent over time, forming segments in offline aggregated scans or overlaid online scans. 
    \newline\mbox{}\hspace*{12pt} 
    In this sample, which features a dynamic scene, both TARL-Seg and 4D-Seg fail to catch the moving vehicle as a single object. On static objects (stopped cars), TARL-Seg is a bit better than 4D-Seg. Yet, for most such objects, 4D-Seg catches with a single ID the majority of the object points. It provides enough regularization, after training \ours (which is online) on offline 4D-Seg data, so that, for all static cars, points of an object instance mostly get a single ID. The moving car is still split into several IDs, but considerably less than in TARL-Seg or 4D-Seg. In fact, it seems that with \ours, the car is more or less consistently tracked over time, but is split into different parts.}
\end{figure*}

\begin{figure*}[p]
    \centering 
    \begin{minipage}{0.48\linewidth}
    \includegraphics[width=\linewidth]{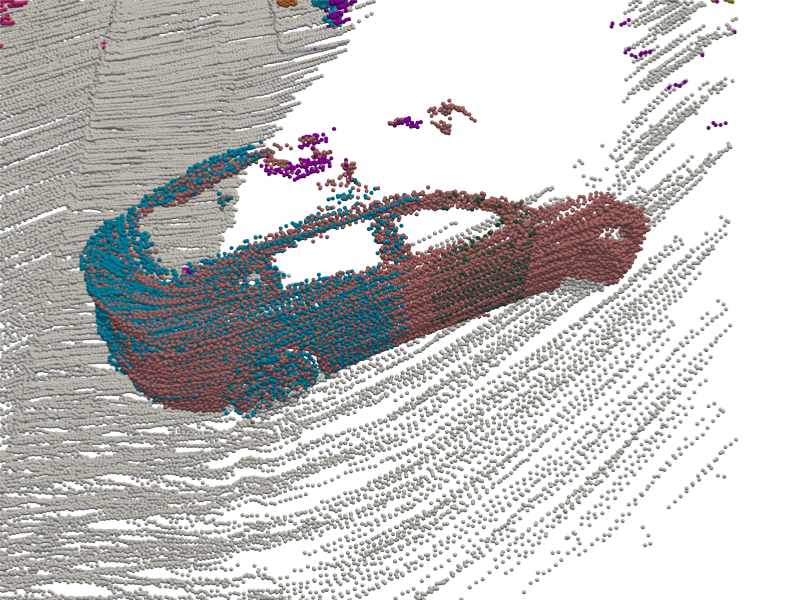}
    \centering \small TARL-Seg: offline 4D segments from TARL computed on \\ aggregated scans
    \end{minipage}~~~
    \begin{minipage}{0.48\linewidth}
    \includegraphics[width=\linewidth]{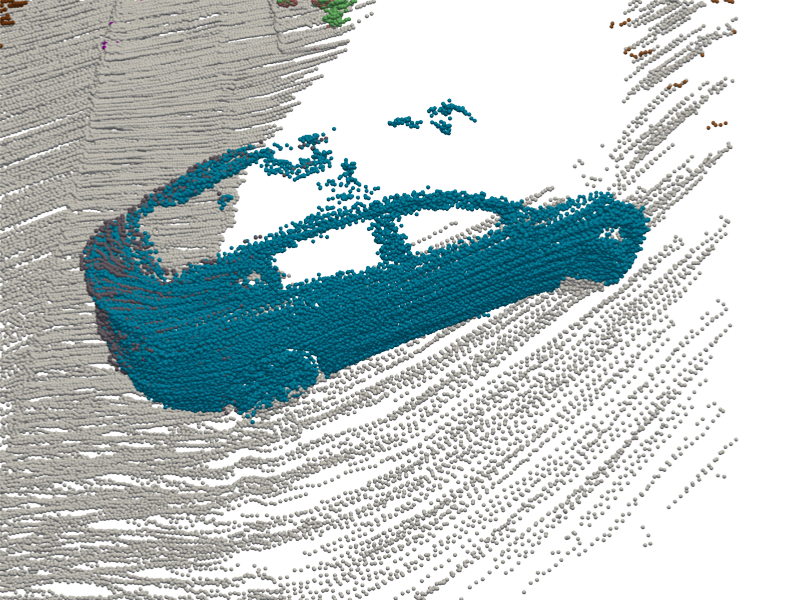}
    \centering \small 4D-Seg: our offline 4D segments computed on \\ aggregated scans
    \end{minipage}
    \\ $\vphantom{x}$ \\
    \begin{minipage}{0.48\linewidth}
    \includegraphics[width=\linewidth]{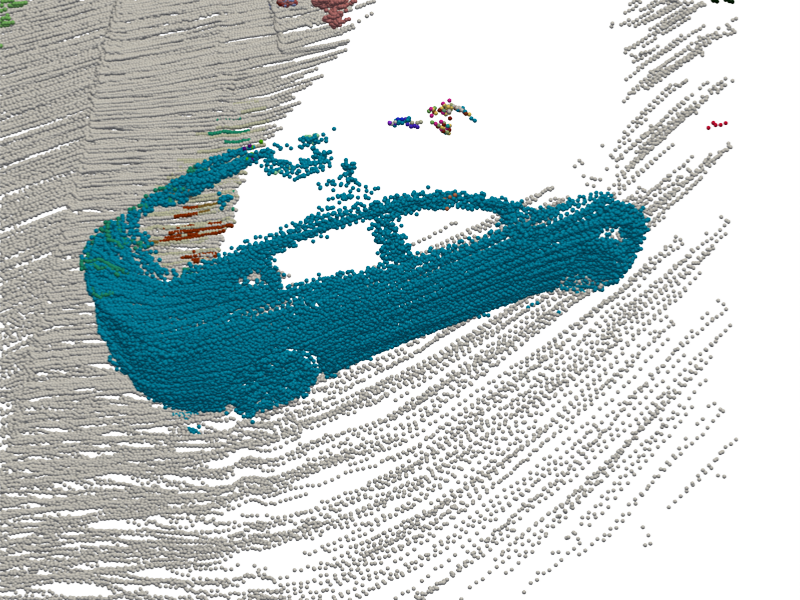}
    \centering \small \ours (trained on 4D-Seg): overlay of our successive \\ online single-scan segments
    \end{minipage}~~~
    \begin{minipage}{0.48\linewidth}
    \includegraphics[width=\linewidth]{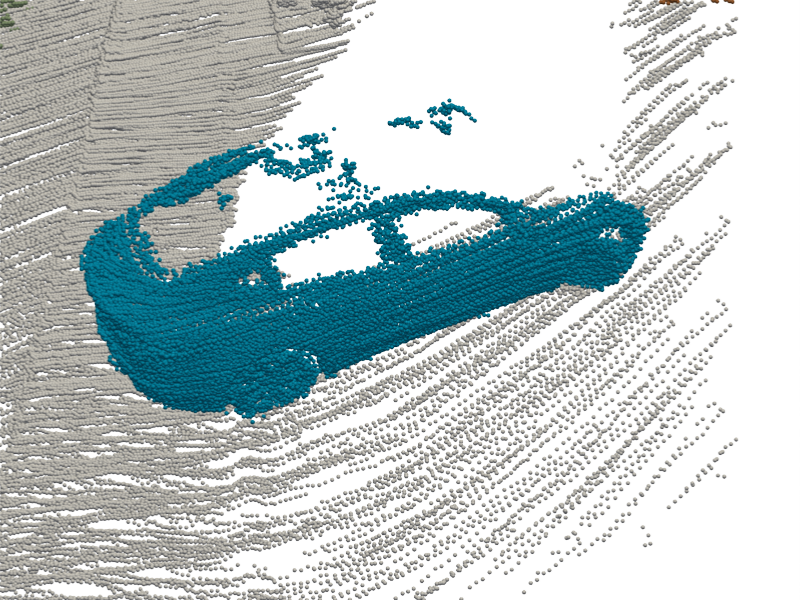}
    \centering \small GT: overlay of ground-truth single-scan segments \\ $\vphantom{\text{Pg}}$
    \end{minipage}
    \caption{Visualization of object instances across time, obtained on a sample scene of PandaSet-GT. Different instances are assigned colors that are random but (tentatively) consistent over time, forming segments in offline aggregated scans or overlaid online scans. 
    \newline\mbox{}\hspace*{12pt} 
    In this sample, which features a parked car, UNIT shows the highest precision of instance segmentation, with 4D-Seg showing very close results albeit. TARL-Seg shows the car instance divided into multiple IDs, showing the limits of its small context window.}
\end{figure*}

\begin{figure*}[p]
    \centering 
    \begin{minipage}{0.48\linewidth}
    \includegraphics[width=\linewidth]{figures/renderings/pd1/SEGTARL.png}
    \centering \small TARL-Seg: offline 4D segments from TARL computed on \\ aggregated scans
    \end{minipage}~~~
    \begin{minipage}{0.48\linewidth}
    \includegraphics[width=\linewidth]{figures/renderings/pd1/Seg4D.png}
    \centering \small 4D-Seg: our offline 4D segments computed on \\ aggregated scans
    \end{minipage}
    \\ $\vphantom{x}$ \\
    \begin{minipage}{0.48\linewidth}
    \includegraphics[width=\linewidth]{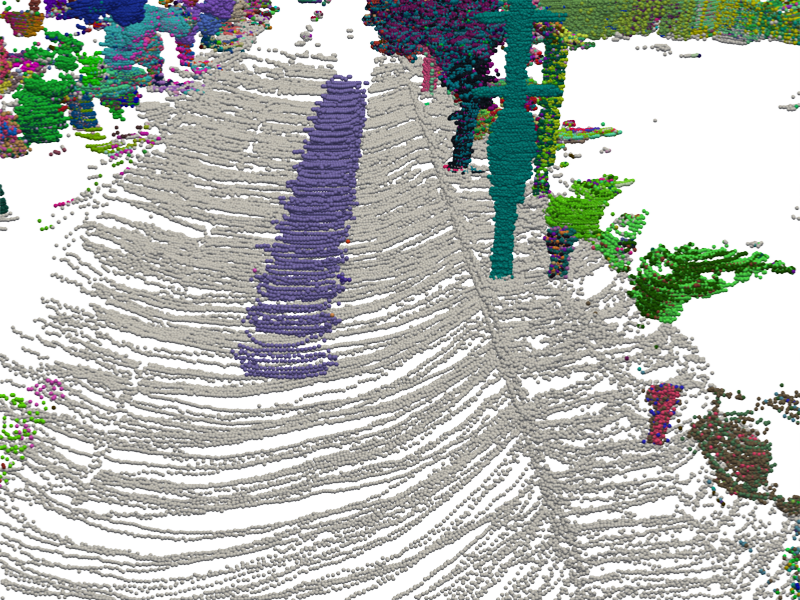}
    \centering \small \ours (trained on 4D-Seg): overlay of our successive \\ online single-scan segments
    \end{minipage}~~~
    \begin{minipage}{0.48\linewidth}
    \includegraphics[width=\linewidth]{figures/renderings/pd1/GT.png}
    \centering \small GT: overlay of ground-truth single-scan segments \\ $\vphantom{\text{Pg}}$
    \end{minipage}
    \caption{Visualization of object instances across time, obtained on a sample scene of PandaSet-GT (cf. Fig.\,1). Different instances are assigned colors that are random but (tentatively) consistent over time, forming segments in offline aggregated scans or overlaid online scans. 
    \newline\mbox{}\hspace*{12pt} 
    In this sample, which features a dynamic scene, TARL-Seg and 4D-Seg have a similar qualitative performance: the background stuff and objects are a bit noisy while the moving car is segmented with two IDs instead of one.
    After training \ours (which is online) on offline 4D-Seg data, a regularization operates and the moving car is now assigned a single ID over time, apart from just a few noisy points. 
    \newline\mbox{}\hspace*{12pt}
    Please note that we (4D-Seg and \ours) obtain more labels than in the ground truth as our class-agnostic segmentation include all objects and stuff, such as trees or buildings, while the ground truth is restricted to a few selected object classes.}
\end{figure*}

\end{document}